\documentclass[10pt,twocolumn,letterpaper]{article}

\usepackage{cvpr}
\usepackage{times}
\usepackage{epsfig}
\usepackage{graphicx}
\usepackage{amsmath}
\usepackage{amssymb}
\usepackage{listings}
\usepackage{diagbox}
\usepackage{amsthm}
\usepackage{amsmath}
\usepackage{amssymb}
\usepackage{pbox}
\usepackage{url}

\usepackage{algorithm}
\usepackage{algpseudocode}

\usepackage{epsfig}
\usepackage{graphicx}
\usepackage{subcaption}
\usepackage{enumerate}

\usepackage[misc]{ifsym}
\usepackage{bbding}

\usepackage{booktabs}
\usepackage{pifont}
\newcommand{\cmark}{\ding{51}}%
\newcommand{\xmark}{\ding{55}}%

\newcommand{\ba}{\mathbf{a}}
\newcommand{\bc}{\mathbf{c}}

\newcommand{\bs}{\mathbf{s}}
\newcommand{\bb}{\mathbf{b}}
\newcommand{\bp}{\mathbf{p}}
\newcommand{\hbb}{\hat{\bb}}
\newcommand{\hbc}{\hat{\bc}}
\newcommand{\hba}{\hat{\ba}}

\newcommand{\hbs}{\hat{\bs}}

\newcommand{\hbp}{\hat{\bp}}

\usepackage[table,x11names]{xcolor}

\usepackage{multicol}
\usepackage{multirow}

\newcommand\paperurl[1]{{\footnotesize{\color{blue}{\url{#1}}}}}

\newcommand{\comment}[1]{}

\definecolor{LightCyan}{rgb}{0.88,1,1}

\newlength\savewidth\newcommand\shline{\noalign{\global\savewidth\arrayrulewidth
  \global\arrayrulewidth 1pt}\hline\noalign{\global\arrayrulewidth\savewidth}}

\makeatletter
\def\@fnsymbol#1{\ensuremath{\ifcase#1\or \S\or \S\or
\mathsection\or \mathparagraph\or \|\or **\or \S\S
\or \ddagger\ddagger \else\@ctrerr\fi}}
\makeatother

\newcommand{\tablestyle}[2]{\ttfamily\setlength{\tabcolsep}{#1}\renewcommand{\arraystretch}{#2}\centering\footnotesize}

\usepackage[pagebackref=true,breaklinks=true,letterpaper=true,colorlinks,bookmarks=false]{hyperref}

\usepackage{algorithm}
\usepackage[]{algpseudocode}

\usepackage[capitalize]{cleveref}
\crefname{section}{Sec.}{Secs.}
\Crefname{section}{Section}{Sections}
\Crefname{table}{Table}{Tables}
\crefname{table}{Tab.}{Tabs.}

\cvprfinalcopy 

\begin{document}

\title{\bf{V-DETR: DETR with Vertex Relative Position Encoding for 3D Object Detection}}

\author{
Yichao Shen$^{1}{\thanks{Core contribution. \Letter\; \texttt{yuhui.yuan@microsoft.com}}}$\quad\;
Zigang Geng$^{2\S}$ \quad
Yuhui Yuan$^{3\S}$ \quad
Yutong Lin$^{1}$ \quad
Ze Liu$^{2}$ \quad \\
Chunyu Wang$^3$ \quad\quad\;
Han Hu$^3$ \quad\quad\;
Nanning Zheng$^{1}$ \quad\quad\;
Baining Guo$^{3}$ \\\hspace{-5mm}
$^1$Xi'an Jiaotong University \quad
$^2$University of Science and Technology of China \quad
$^3$Microsoft Research Asia
}

\maketitle

\begin{abstract}
We introduce a highly performant 3D object detector for point clouds using the DETR framework. The prior attempts all end up with suboptimal results because they fail to learn accurate inductive biases from the limited scale of training data. In particular, the queries often attend to points that are far away from the target objects, violating the locality principle in object detection. To address the limitation, we introduce a novel 3D Vertex Relative Position Encoding (3DV-RPE) method which computes position encoding for each point based on its relative position to the 3D boxes predicted by the queries in each decoder layer, thus providing clear information to guide the model to focus on points near the objects, in accordance with the principle of locality. In addition, we systematically improve the pipeline from various aspects such as data normalization based on our understanding of the task. We show exceptional results on the challenging ScanNetV2 benchmark, achieving significant improvements over the previous 3DETR in $\rm{AP}_{25}$/$\rm{AP}_{50}$ from 65.0\%/47.0\% to 77.8\%/66.0\%, respectively. In addition, our method sets a new record on ScanNetV2 and SUN RGB-D datasets.
Code will be released at: {\url{{https://github.com/yichaoshen-MS/V-DETR}}}.
\end{abstract}

\begin{figure}[t]
\centering
\includegraphics[height=0.2\columnwidth]{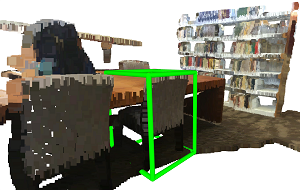}
\hfill
\includegraphics[height=0.2\columnwidth]{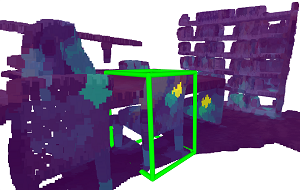}
\hfill
\includegraphics[height=0.2\columnwidth]{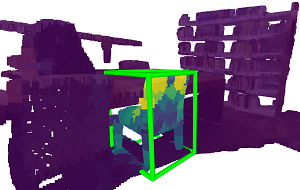} \\
\includegraphics[height=0.2\columnwidth]{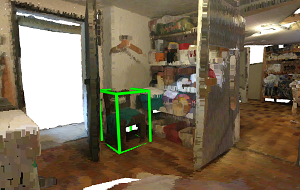}
\hfill
\includegraphics[height=0.2\columnwidth]{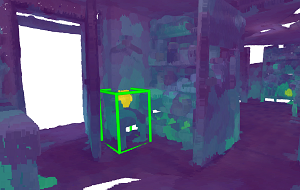}
\hfill
\includegraphics[height=0.2\columnwidth]{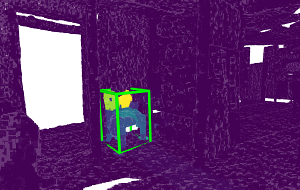} \\
\begin{subfigure}[b]{0.3\linewidth}
\centering
\includegraphics[height=0.64\columnwidth]{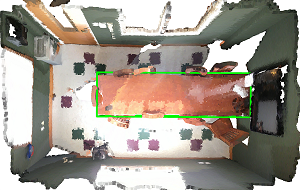}
\caption{}
\end{subfigure}
\hfill
\begin{subfigure}[b]{0.3\linewidth}
\centering
\includegraphics[height=0.64\columnwidth]{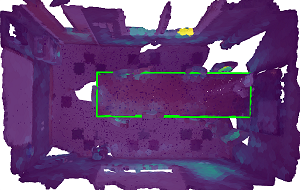}
\caption{}
\end{subfigure}
\hfill
\begin{subfigure}[b]{0.3\linewidth}
\centering
\includegraphics[height=0.64\columnwidth]{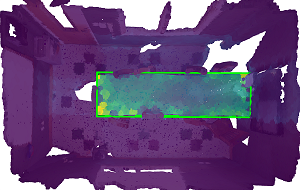} 
\caption{}
\end{subfigure}

\caption{\small{
(a) 3D scans from the ScanNetV2 in the rear/front/top-down view. 
We display one of the ground-truth bounding boxes with a green 3D box.
(b) The decoder cross-attention map based on plain DETR. Attention weights are distributed over many positions even outside the ground-truth box.
(c) The decoder cross-attention map based on plain DETR + 3DV-RPE. Attention weights focus on the sparse object boundaries of the object located in the ground-truth bounding box.
The color indicates the attention values: yellow for high and blue for low.
}}
\label{fig:attention_maps_intro}
\vspace{-5mm}
\end{figure}

\section{Introduction}
\begin{figure*}[t]
\centering
\hspace{-12mm}
\begin{subfigure}[b]{0.22\linewidth}
\centering
\includegraphics[height=0.32\columnwidth]{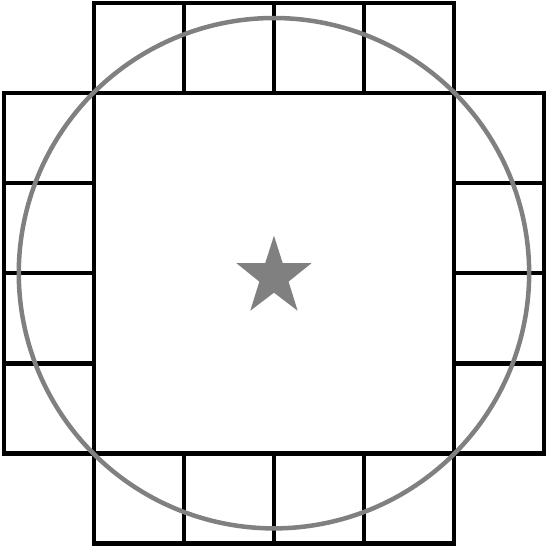}
\caption{\small{voxelized sparse input}}
\end{subfigure}
\hspace{-2mm}
\begin{subfigure}[b]{0.22\linewidth}
\centering
\includegraphics[height=0.32\columnwidth]{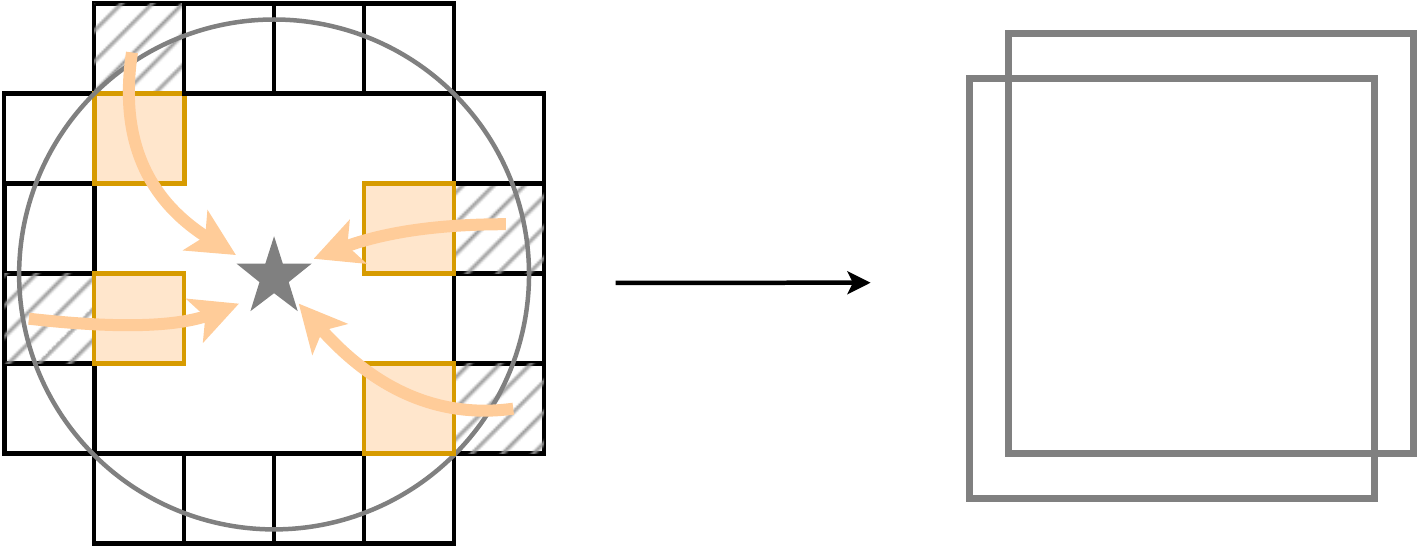}
\caption{\small{Voting-based method}}
\end{subfigure}%
\hspace{5mm}
\begin{subfigure}[b]{0.22\linewidth}
\centering
\includegraphics[height=0.32\columnwidth]{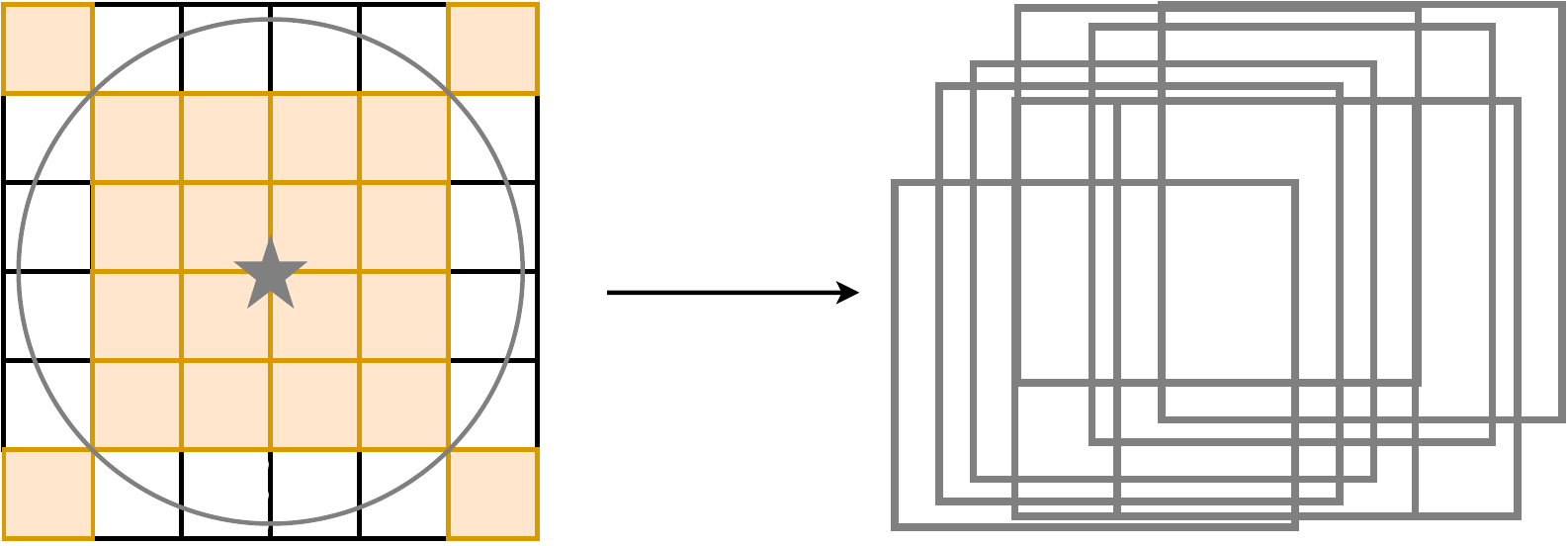}
\caption{\small{Expansion-based method}}
\end{subfigure}%
\hspace{10mm}
\begin{subfigure}[b]{0.22\linewidth}
\centering
\includegraphics[height=0.32\columnwidth]{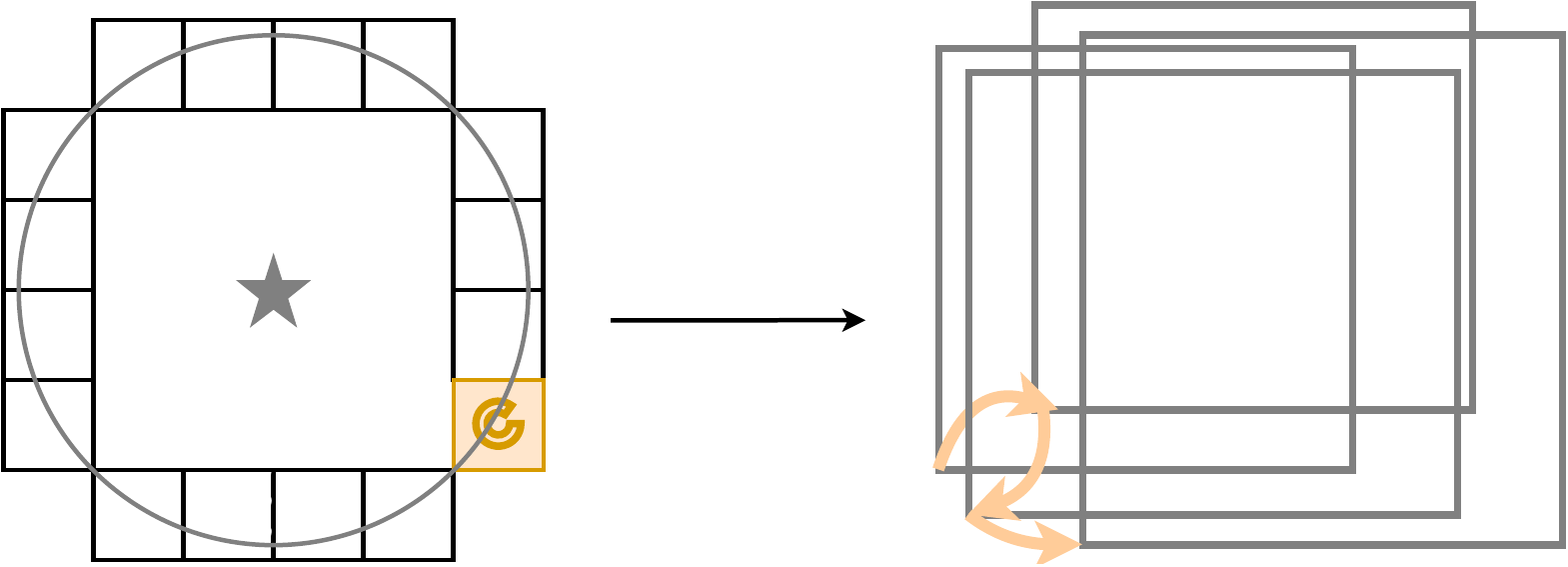}
\caption{DETR-based method}
\end{subfigure}%
\caption{\small{
(a) A simplified sparse 3D voxel space from a top-down perspective. The curve shows the input surface and the small cubes show the voxelized input. The gray five-pointed star \FiveStar shows the object's center. (b) The voting scheme estimates offsets for each voxel and we color the voxels (nearer to the object's center) with yellow after voting. The dashed small cubes show the empty space after voting. (c) The generative sparse decoder (GSD) scheme enlarges the voxels around the surfaces, thus creating new voxels both inside and outside of the object (marked with yellow cubes).
(d) The DETR-based approach simply selects a small set of voxels (marked with yellow cubes) as the initial object query and iteratively predicts the boxes by refining (marked with the open yellow circles) the object query with multiple Transformer decoder layers. We follow the DETR-based path in this work.
}}
\label{fig:intro_compare_approach}
\vspace{-5mm}
\end{figure*}

3D object detection from point clouds is a challenging task that involves identifying and localizing the objects of interest present in a 3D space. This space is represented using a collection of data points that have been gleaned from the surfaces of all accessible objects and background in the scene.  The task has significant implications for various industries, including augmented reality, gaming, robotics, and autonomous driving.

Transformers have made remarkable advancement in 2D object detection, serving as both powerful backbones~\cite{Vaswani2017attention,liu2021swin} and detection architectures~\cite{carion2020end}. However, their performance in 3D detection~\cite{misra2021-3detr} is significantly worse than the state-of-the-art methods.  Our in-depth evaluation of~\cite{misra2021-3detr} revealed that the queries often attend to points that are far away from the target objects (Figure \ref{fig:attention_maps_intro} (b) shows three typical visualizations), which violates the principle of \emph{locality} in object detection. The principle of locality dictates that object detection should only consider subregions of data that contain the object of interest and not the entire space. 
Besides, the behavior is also in contrast with the success that Transformers have achieved in 2D detection, where they have been able to effectively learn the inductive biases, including locality. We attribute the discrepancy to the limited scale of training data available for 3D object detection, making it difficult for Transformers to acquire the correct inductive biases. 

In this paper, we present a simple yet highly performant method for 3D object detection using the transformer architecture DETR~\cite{carion2020end}. To improve locality in the cross-attention mechanism, we introduce a novel 3D Vertex Relative Position Encoding (3DV-RPE) method. It computes a position encoding for each point based on its relative offsets to the vertices of the predicted 3D boxes associated with the queries, providing clear positional information such as whether each point is inside the boxes. This information can be utilized by the model to guide cross-attention to focus on points inside the box, in accordance with the principle of locality. The prediction of these boxes is consistently refined as the decoder layers progress, resulting in increasingly accurate position encoding.

To mitigate the impact of object rotation, we propose to compute 3DV-RPE in a canonical object space where all objects are consistently rotated. Particularly, for each query, we predict a rotated 3D box and compute the relative offsets between the 3D points rotated in the same way, and the eight vertices of the box.
This results in consistent position encoding for different instances of the same object regardless of their positions or orientations in the space, greatly facilitating the learning of the locality property in cross-attention even from limited training data. Figure \ref{fig:attention_maps_intro} (c) visualizes the attention weights obtained by our method. We can see that the query for detecting the chair nicely attends to the points on the chair. Our experiment demonstrates that 3DV-RPE boosts the performance.

We also systematically enhance our pipeline from various aspects such as data normalization and network architectures based on our understanding of the task. For example, we propose object-based normalization, instead of the scene-based one used by the DETR series, to parameterize the 3D boxes. This is because the former is more stable for point clouds which differs from 2D detection where the sizes of the same object in images can be very different depending on the camera parameters, impelling them to use image size to coarsely normalize the boxes. Besides, we also evaluate and adapt some of the recent advancement in 2D DETR.

We conduct thorough experiments to empirically show that our simple DETR-based approach significantly outperforms the previous state-of-the-art fully convolutional 3D detection methods,
which helps to accelerate the convergence of the detection head architecture design for 2D and 3D detection tasks.
We report the results of our approach on two challenging indoor 3D object detection benchmarks including ScanNetV$2$ and SUN RGB-D. Overall, compared to the DETR baseline~\cite{misra2021-3detr}, our method with 3DV-RPE improves $\rm{AP}_{25}$/$\rm{AP}_{50}$ from $65.0\%$/$47.0\%$ to $77.8\%$/$66.0\%$, respectively, and reduces the training epochs by 50\%.
Particularly, on ScanNetV$2$, our approach outperforms the very recent state-of-the-art CAGroup3D~\cite{wang2022cagroup3d} by +$2.7\%$/+$4.7\%$ measured by $\rm{AP}_{25}$/$\rm{AP}_{50}$, respectively.

\section{Related work}

\vspace{1mm}
\noindent\textbf{DETR-based Object Detection.}
DETR~\cite{carion2020end} is a groundbreaking work that applies transformers~\cite{Vaswani2017attention} to $2$D object detection, eliminating many hand-designed components such as non-maximum suppression~\cite{neubeck2006efficient} or anchor boxes~\cite{girshick2015fast,ren2015faster,lin2017focal,liu2016ssd}. Many extensions of DETR have been proposed~\cite{meng2021CondDETR,gao2021fast,dai2021dynamic,wang2021anchor,jia2022detrs,zhang2022dino}, such as Deformable-DETR~\cite{zhu2020deformable}, which uses multi-scale deformable attention to focus on key sampling points and improve performance on small objects. DAB-DETR~\cite{liu2022dab} introduces a novel query formulation to enhance detection accuracy. Some recent works~\cite{li2022dn,zhang2022dino,jia2022detrs,chen2022group} achieve state-of-the-art results on object detection by using query denoising or one-to-many matching schemes, which addressed the training inefficiency of one-to-one matching. $\mathcal{H}$-DETR~\cite{jia2022detrs} shows that one-to-many matching can also speed up convergence on 3D object detection tasks. Following the DETR-based approach, GroupFree~\cite{liu2021group} and 3DETR~\cite{misra2021-3detr} built strong 3D object detection systems for indoor scenes. However, they are still inferior to other methods such as CAGroup3D~\cite{wang2022cagroup3d}. In this work, we propose several critical modifications to improve the DETR-based methods and achieve new records on two indoor 3D object detection tasks.

\vspace{1mm}
\noindent\textbf{3D Indoor Object Detection.}
We revisit the existing indoor 3D object detection methods that directly use raw point clouds to detect 3D boxes. We categorize them into three types based on their strategies:
(i) \emph{Voting-based methods}, such as VoteNet~\cite{qi2019deep}, MLCVNet~\cite{xie2020mlcvnet} and H3DNet~\cite{zhang2020h3dnet}, use a voting mechanism to shift the surface points toward the object centers and group them into object candidates.
(ii) \emph{Expansion-based methods}, such as GSDN\cite{gwak2020generative}, FCAF$3$D\cite{rukhovich2022fcaf3d}, and CAGroup$3$D\cite{wang2022cagroup3d}, which generate virtual center features from surface features using a generative sparse decoder and predict high-quality 3D region proposals.
(iii) \emph{DETR-based methods},
unlike these two types that require modifying the original geometry structure of the input 3D point cloud, we adopt the DETR-based approach~\cite{liu2021group,misra2021-3detr} for its simplicity and generalization ability. Our experiments show that DETR has great potential for indoor 3D object detection.
We show the differences between above-mentioned methods in Figure~\ref{fig:intro_compare_approach}.

\vspace{1mm}
\noindent\textbf{3D Outdoor Object Detection.}
We briefly review some methods for outdoor 3D object detection~\cite{yan2018second,zhou2018voxelnet,lang2019pointpillars,yin2021center}, which mostly transform 3D points into a bird-eye-view plane and apply 2D object detection techniques. For example, VoxelNet~\cite{zhou2018voxelnet} is a single-stage and end-to-end network that combines feature extraction and bounding box prediction. PointPillars~\cite{lang2019pointpillars} uses a 2D convolution neural network to process the flattened pillar features from a Bird’s Eye View (BEV). CenterPoint~\cite{yin2021center} first detects centers of objects using a keypoint detector and regresses to other attributes, then refines them using additional point features on the object. However, these methods still suffer from center feature missing issues, which FSD~\cite{fan2023super} tries to address. We plan to extend our approach to outdoor 3D object detection in the future, which could unify indoor and outdoor 3D detection tasks.

\begin{figure}[t]
\centering
\includegraphics[width=0.49\textwidth]{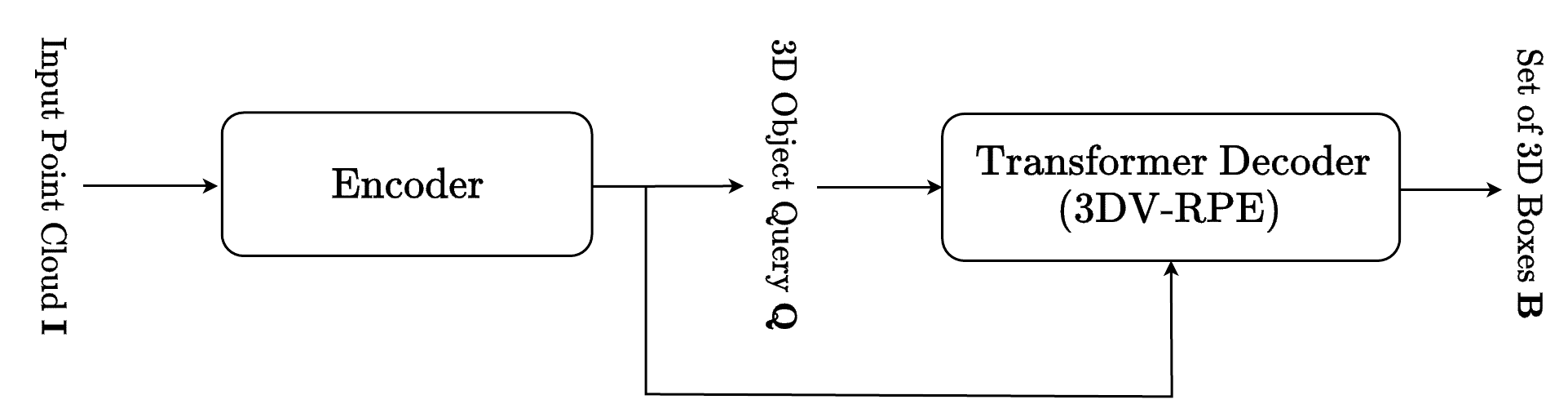}
\caption{\small{\textbf{Illustrating the overall framework of V-DETR for 3D object detection}. We first use an encoder to extract 3D features and then we use a plain Transformer decoder to estimate the 3D object queries from a set of initialized 3D object queries. In the Transformer decoder multi-head cross-attention layer, we use a 3D vertex relative position encoding scheme for both locality and accurate position modeling.}}
\label{fig:vdetr_pipeline}
\end{figure}

\section{Our Approach}

\subsection{Baseline setup}

\vspace{1mm}
\noindent\textbf{Pipeline.}
We build our V-DETR baseline following the previous DETR-based 3D object detection methods~\cite{misra2021-3detr,liu2021group}.
The detailed steps are as follows: given a 3D point cloud $\mathbf{I} {\in} \mathbb{R}^{\mathsf{N}\times \mathsf{6}}$ sampled from a 3D scan of an indoor scene, where the RGB values are in the first $3$ dimensions and the position XYZ values are in the last $3$ dimensions.
We first sample about {$\sim$}$40$K points from the original point cloud that typically has around {$\sim$}$200$K points.
Second, we use a feature encoder to process the raw sampled points and compute the point features $\mathbf{F} {\in} \mathbb{R}^{\mathsf{M}\times \mathsf{C}}$.
Third, we construct a set of 3D object queries $\mathbf{Q} {\in} \mathbb{R}^{\mathsf{K}\times \mathsf{C}}$ send them into a plain Transformer decoder to predict a set of 3D bounding boxes $\mathbf{B} {\in} \mathbb{R}^{\mathsf{K}\times \mathsf{D}}$.
We set $\mathsf{K}=1024$ by default.
Figure~\ref{fig:vdetr_pipeline} shows the overall pipeline.
We present more details on the encoder architecture design, the 3D object query construction, the Hungarian matching, and loss function formulations as follows.

\vspace{1mm}
\noindent\textbf{Encoder architecture.}
We choose two different kinds of encoder architecture for experiments including:
(i) a PointNet followed by a shallow Transformer encoder adopted by~\cite{misra2021-3detr} or (ii) a sparse 3D modification of ResNet$34$ followed by an FPN neck adopted by~\cite{rukhovich2022fcaf3d}, where we replace the expensive generative transposed convolution with a simple transposed convolution within the FPN neck.

\vspace{1mm}
\noindent\textbf{3D object query.}
We construct the 3D object query by combining two kinds of representations as follows: first, we simply sample a set of $\mathsf{K}$ initial center positions over the entire encoder output space and select their representations to initialize a set of 3D content query $\mathbf{Q}_c$. Then we use their XYZ coordinates in the input point cloud space to compute the 3D position query $\mathbf{Q}_p$ with a simple $\operatorname{MLP}$ consisting of two linear layers.
We build the 3D object query by adding the 3D position query to the 3D content query.

\vspace{1mm}
\noindent\textbf{Hungarian matching and loss function.}
We choose the weighted combination of six terms including the bounding box localization regression loss, angular classification and regression loss, and semantic classification loss as the final matching cost functions and training loss functions.
We illustrate the mathematical formulations
as follows:
\begin{equation}
\begin{aligned}
\label{eq.loss}
\mathcal{L}_{\textrm{DETR}} = - \lambda_{1} \mathrm{GIoU}(\hbb, \bb) + \lambda_{2}  \mathcal{L}_\textrm{center}(\hbc, \bc) +\lambda_{3} \mathcal{L}_\textrm{size}(\hbs, \bs)
\notag\\- \lambda_{4} \mathrm{FL}(\hat{\mathbf{p}}[l]) +\lambda_{5} \mathcal{L}_\textrm{huber}(\hba_r, \ba_r) +\lambda_{6} \mathrm{CE}(\hba_c, \ba_c),\notag
\end{aligned}
\end{equation}
where we use $\hbb$, $\hbc$, $\hbs$, $\hbp$, $\hba$ (or $\bb$, $\bc$, $\bs$, $l$, $\ba$) to represent the predicted (or ground-truth) bounding box, box center, box size, classification score, and rotation angle respectively, e.g.,
$l$ represents the ground-truth semantic category of $\bb$.
$\mathrm{CE}$ represents angle classification cross entropy loss and $\mathcal{L}_\textrm{huber}$ represents the residual continuous angle regression loss.
$\mathrm{FL}$ represents semantic classification focal loss.
We ablate the influence of hyper-parameter value choices in the ablation experiments.

\vspace{1mm}
\noindent\textbf{Object-normalized box parameterization.}
We propose an object-normalized box reparameterization scheme that differs from the original DETR~\cite{carion2020end}, which normalizes the box predictions by the scene scales.
We account for one key discrepancy between object size variation in 2D images and 3D point clouds, e.g., a chair's 2D box size may change depending on its distance to the cameras, but its 3D box size should remain consistent as the point cloud captures the real 3D world.
In the implementation, we simply reparameterize the prediction target of width and height from the original groud-truth $\bb_h$ and $\bb_w$ to $\bb_h/\hbb_h^{l-1}$ and $\bb_w/\hbb_w^{l-1}$, where $\hbb_h^{l-1}$ and $\hbb_w^{l-1}$ represent the coarsely predicted box height and width.

\begin{figure}[t]
\centering
\includegraphics[width=0.25\textwidth]{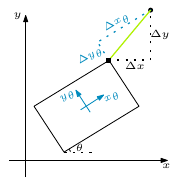}
\caption{\small{\textbf{Canonical object space transformation for 3DV-RPE.}
The rectangle represents the box of the object, which defines an object coordinate system. The green line represents the offset from a point to the box vertex. The offset transformed to the object coordinate system is $(\Delta x_{\theta}, \Delta y_{\theta})$ where the exact values can be geometrically reasoned. Since there is no rotation along the z-axis on the current datasets, we only show the changes in the x-y plane.}}
\label{fig:rotatedRPE}
\vspace{-2mm}
\end{figure}

\vspace{3mm}
\subsection{3DV-RPE in Canonical Object Space}

Position Encoding (PE) is crucial for enhancing the ability of transformers to comprehend the spatial context of the tokens. The appropriate PE strategy depends on tasks. For 3D object detection, where geometry features are the primary focus, it is essential for PE to encode rich semantic positions for the points, \eg whether they are on/off the 3D shapes of interest. 
\vspace{0.5em}

To that end, we present 3D Vertex Relative Position Encoding (3DV-RPE), a novel solution specifically tailored for 3D object detection within the DETR framework. We modify the global plain Transformer decoder multi-head cross-attention maps as follows:
\begin{equation}
\begin{aligned}
\label{eq:modulate_cross_attn}
  \widehat{\mathbf{A}} = \operatorname{Softmax}(\mathbf{Q}\mathbf{K}^{\text{T}}\textcolor{black}{\;+ \;\mathbf{R}}),
\end{aligned}
\end{equation}
where $\mathbf{Q}$ and $\mathbf{K}$ represent the sparse query points and the dense key-value points, respectively.
$\mathbf{R}$ represents the position encoding computed by our 3DV-RPE that carries accurate position information.

\begin{figure}[t]
\centering
\centering
\includegraphics[width=0.48\textwidth]{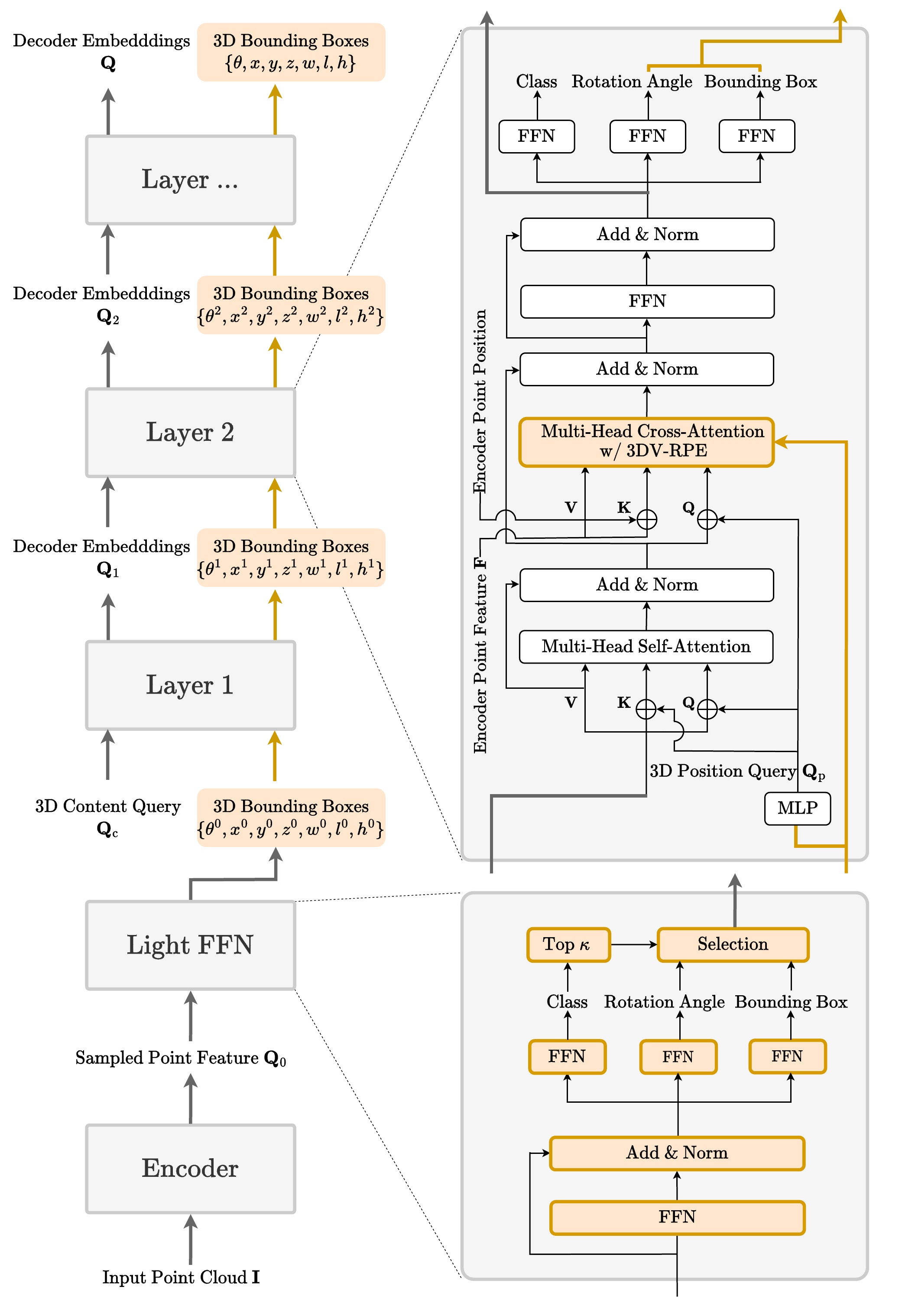}
\caption{\small{\textbf{Illustration of the proposed V-DETR framework.} We mark the modifications with yellow-colored regions and the other components, that are designed following the plain DETR, with gray-colored regions. 
}}
\label{fig:pipeline_details}
\end{figure}

\paragraph{3DV-RPE.}
Our key insight is that, encoding a point by its relative position to the target object, which is coarsely represented by a box, is sufficient for 3D object detection. It is computed as follows:
\begin{equation}
\begin{aligned}
\label{eq.rpe_transform}
{\mathbf{P}_i} = \operatorname{MLP}_i(\mathcal{F}(\Delta\mathbf{P}_{i})), 
\end{aligned}
\end{equation}
where $\Delta\mathbf{P}_{i}\in \mathbb{R}^{\mathsf{K}\times\mathsf{N}\times3}$ denotes the offsets between the $N$ points and the $i$-th vertex of the $K$ boxes and $\mathbf{P}_i\in \mathbb{R}^{\mathsf{K}\times\mathsf{N}\times h}$ represents the relative position bias term. $h$ is the number of heads. $\mathcal{F}(\cdot)$ is a non-linear function. We will evaluate several alternatives for $\mathcal{F}(\cdot)$ in the experiments. $\operatorname{MLP}_i$ represents an MLP based transformation that first projects the features to a higher dimension space, and then to the output features of dimension $h$.

We obtain the final relative position bias term by adding the bias term of the eight vertices, respectively:
\begin{equation}
\begin{aligned}
\label{eq.rpe_add}
{\mathbf{R}} = \sum_{i=1}^{8} \mathbf{P}_i,
\end{aligned}
\end{equation}
where 
${\mathbf{R}}$, encodes the relations between the 3D boxes and the points. In the subsequent section, we will introduce how we compute $\Delta\mathbf{P}_{i}$ with the aid of the boxes predicted at current layer.

\begin{figure*}[t]
\centering
\includegraphics[height=0.32\columnwidth, trim={80 50 80 50},clip]{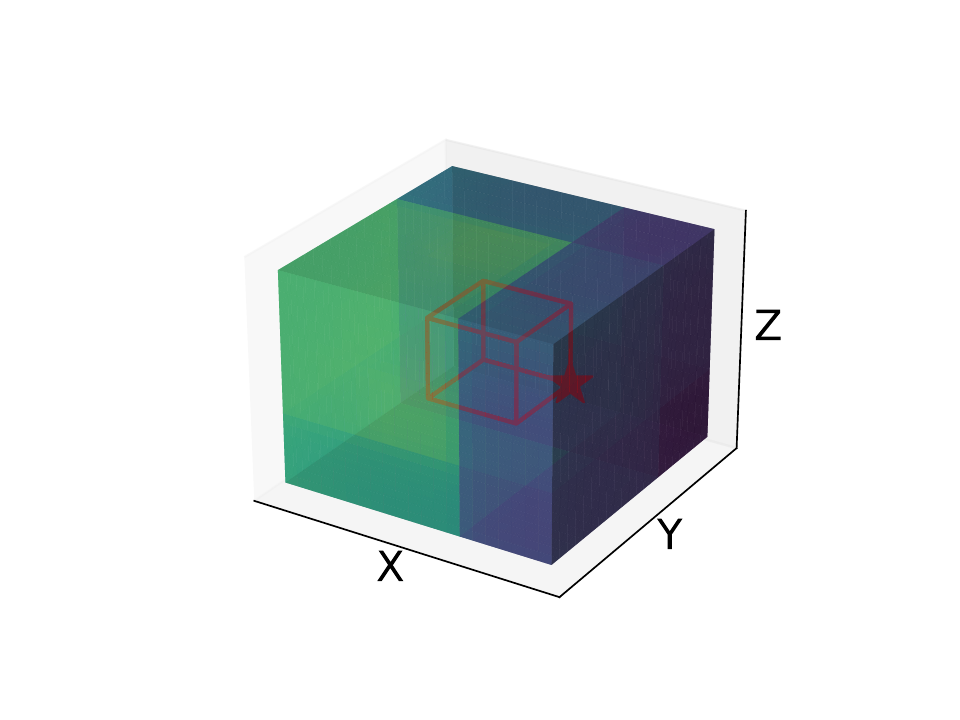}
\includegraphics[height=0.32\columnwidth, trim={80 50 80 50},clip]{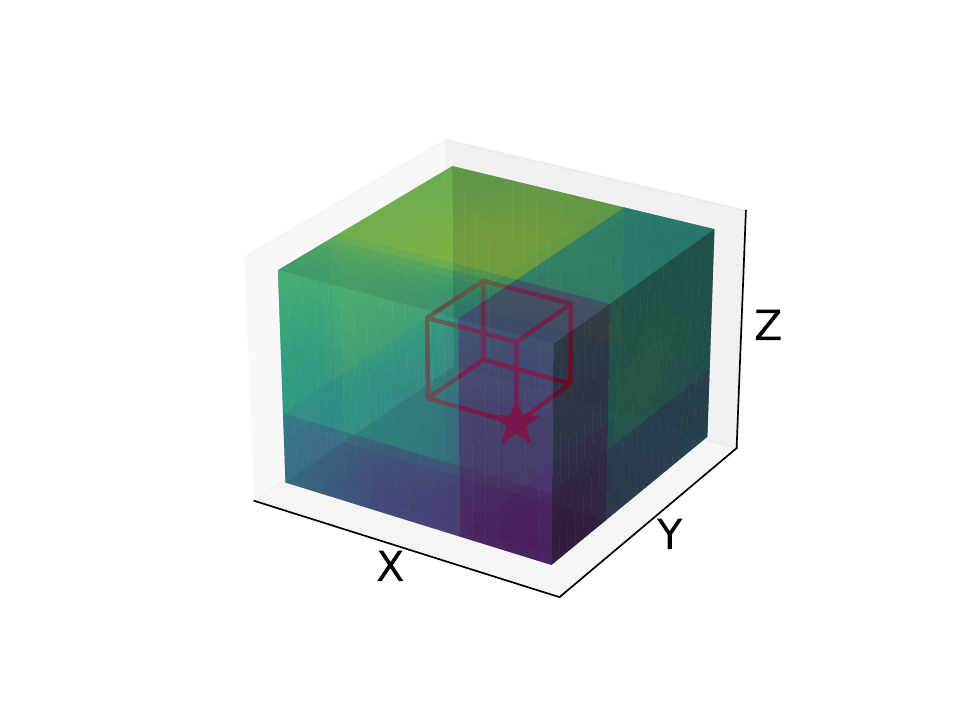}
\includegraphics[height=0.32\columnwidth, trim={80 50 80 50},clip]{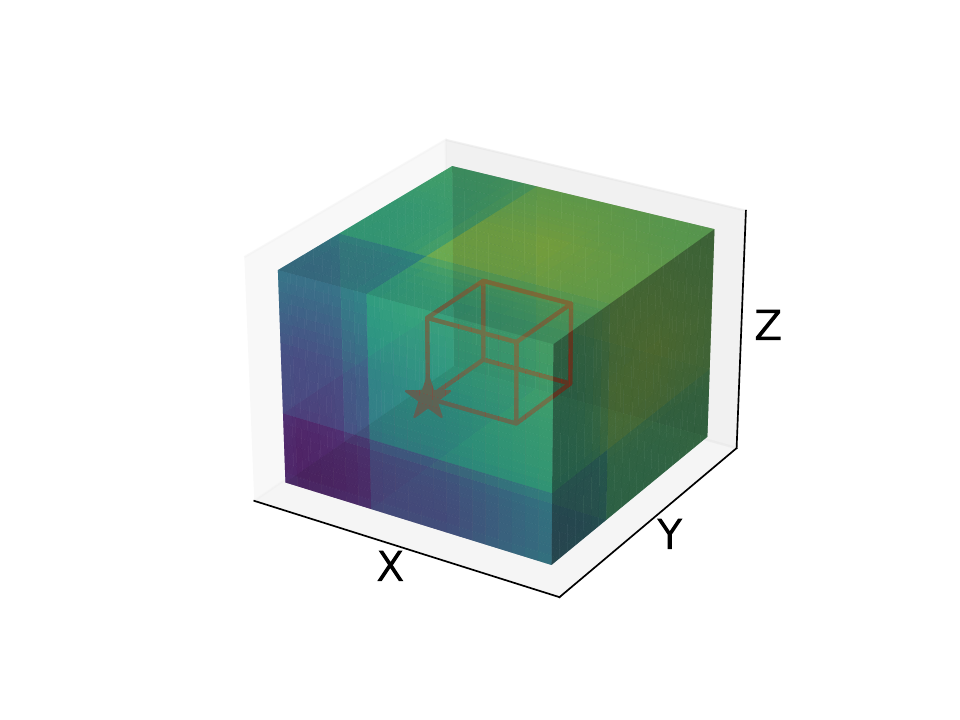}
\includegraphics[height=0.32\columnwidth, trim={80 50 80 50},clip]{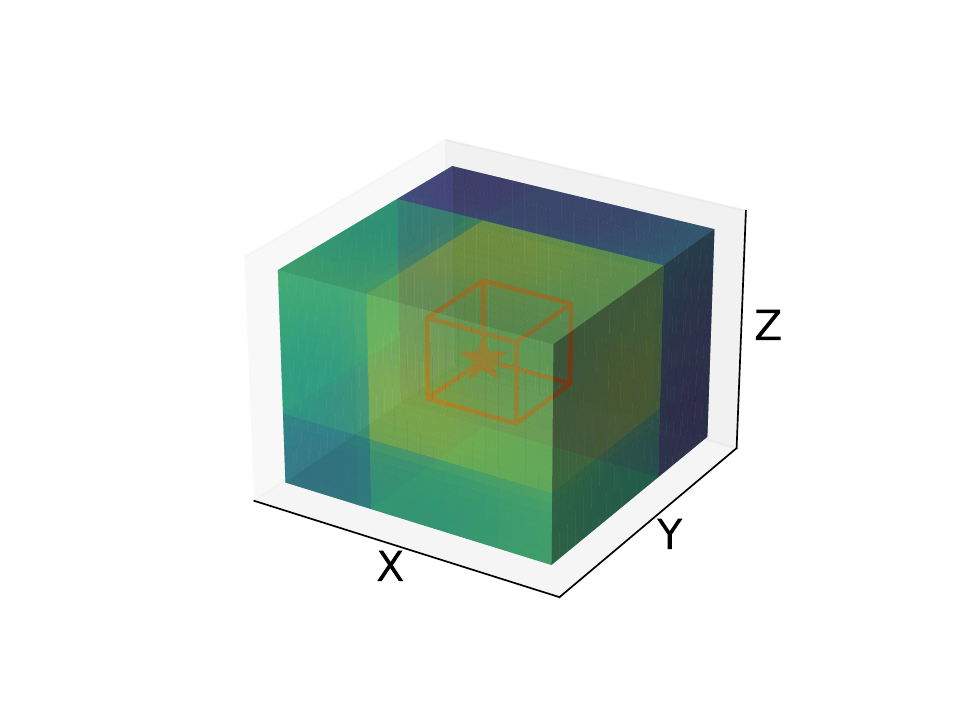}
\includegraphics[height=0.32\columnwidth, trim={80 50 80 50},clip]{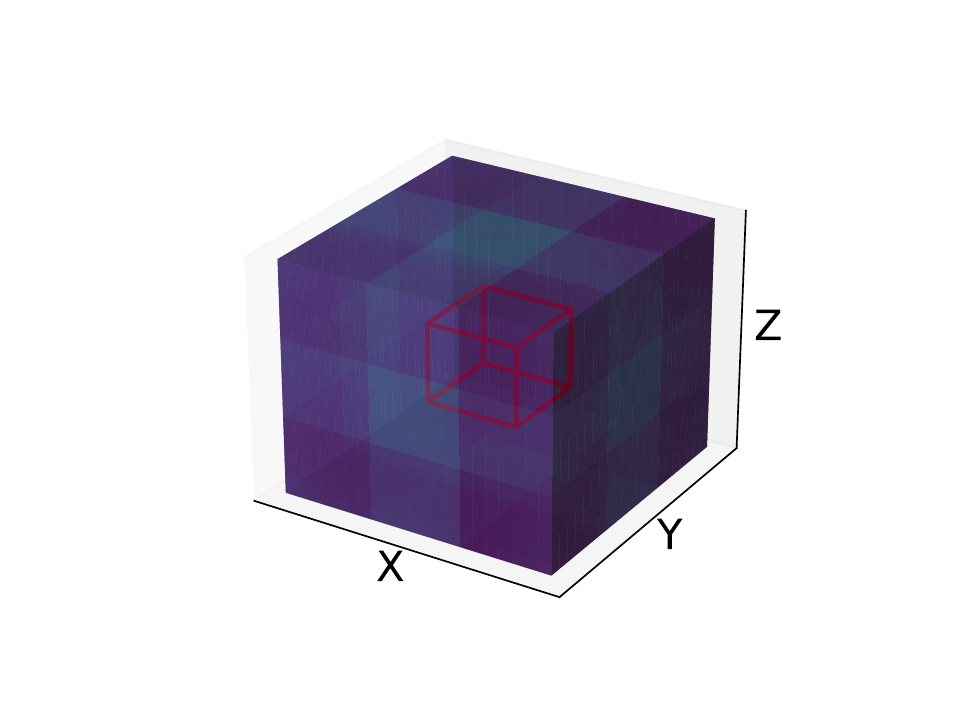}
\includegraphics[height=0.32\columnwidth, trim={80 50 80 50},clip]
{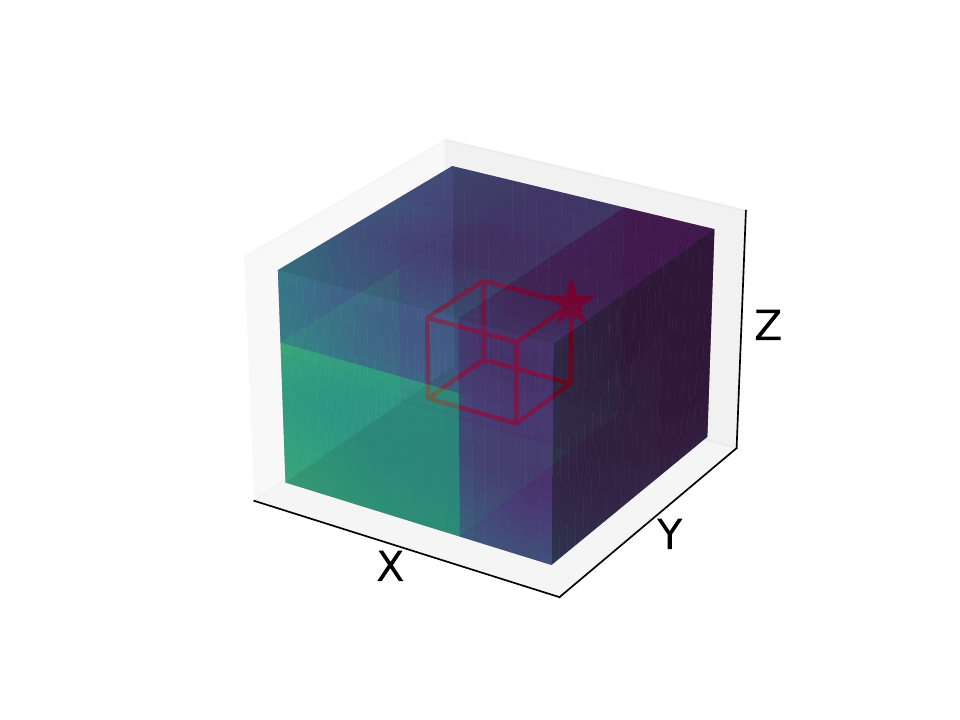}
\includegraphics[height=0.32\columnwidth, trim={80 50 80 50},clip]{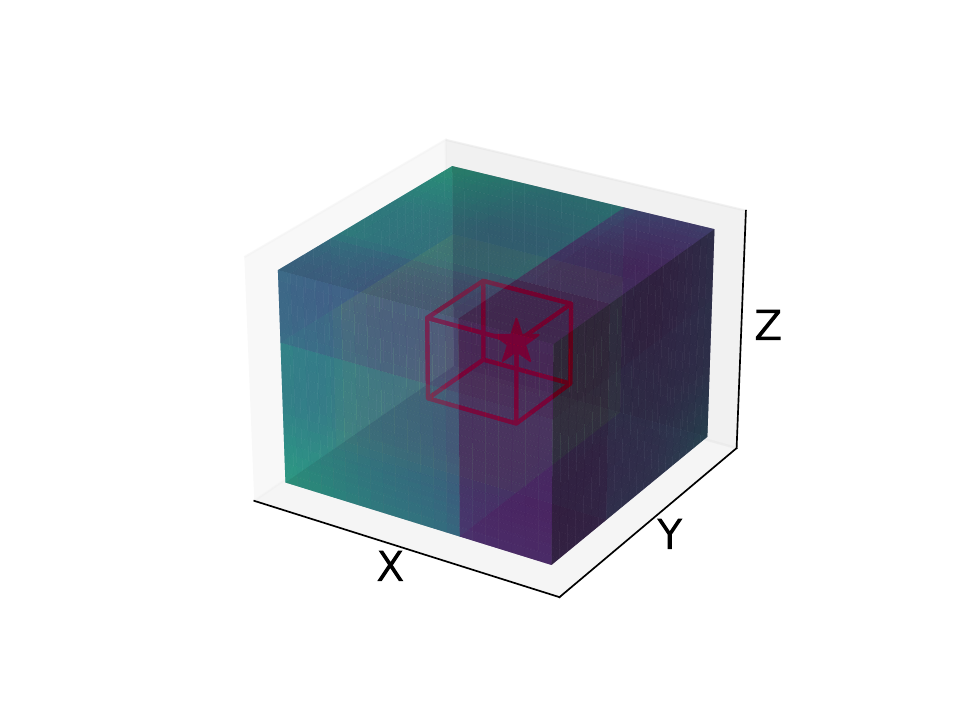}
\includegraphics[height=0.32\columnwidth, trim={80 50 80 50},clip]{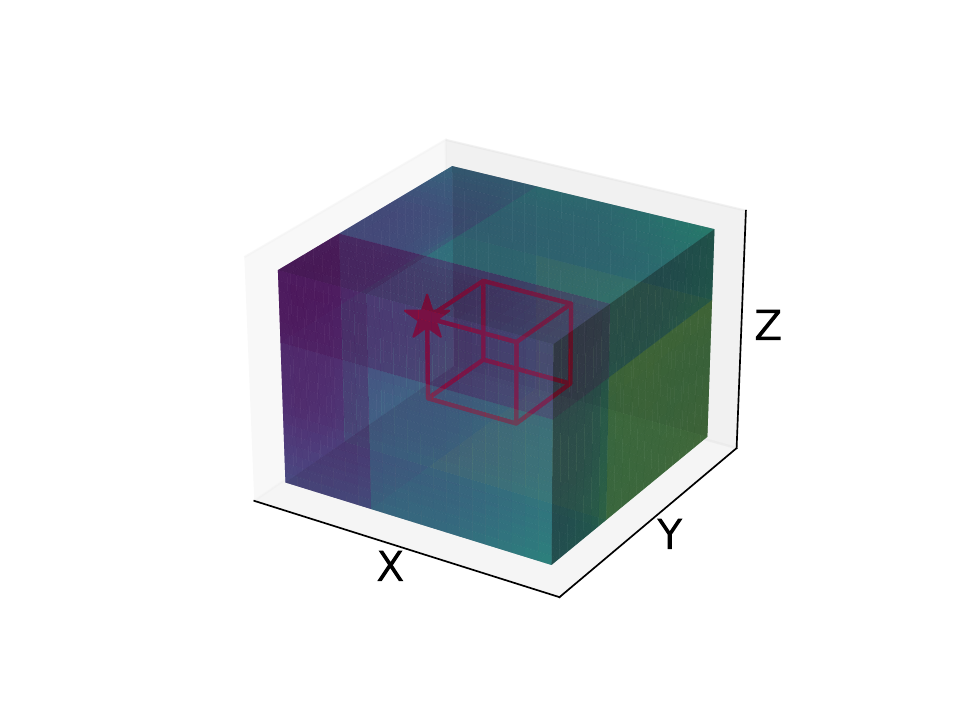}
\includegraphics[height=0.32\columnwidth, trim={80 50 80 50},clip]{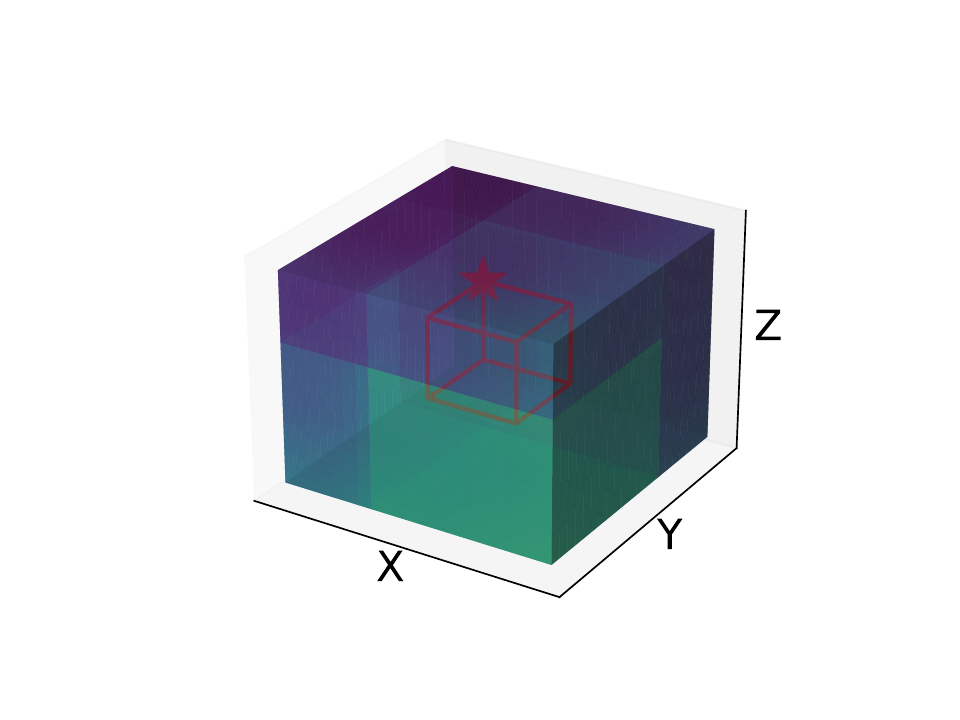}
\includegraphics[height=0.32\columnwidth, trim={80 50 80 50},clip]{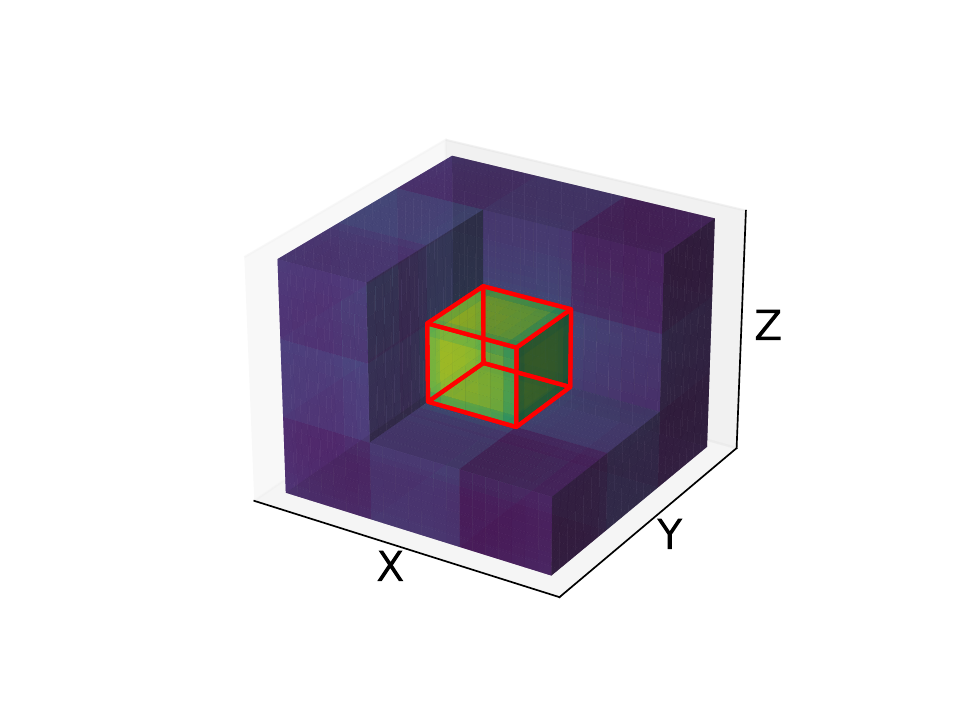}
\caption{\small{\textbf{Visualizing the learned spatial attention maps based on 3DV-RPE.} We use the small red-colored cube to represent the 3D bounding box of an object, the red five-pointed star to mark the eight vertices, and the entire colored cube as the input scene for simplicity. We average each $\mathbf{P}_i$ along head dimension according to Equation~\ref{eq.rpe_transform} and visualize eight vertices' learned spatial cross-attention maps (from column\#1 to column\#4). We visualize the merged spatial attention maps in column\#5 (from the cutaway view). The color indicates the attention values: yellow for high and blue for low. We can observe that (i) the learned spatial attention maps of each vertex can enhance the regions along the internal direction starting from each vertex position, and (ii) the combined spatial attention maps can accurately enhance the internal regions inside the red-colored cubes.}}
\label{fig:attention_maps_rpe}
\vspace{-2mm}
\end{figure*}

\paragraph{Canonical Object Spaces.}
It is worth noting that the direction of the offsets are dependent on the definition of the world coordinate system and the object orientation which complicates the learning of semantic position encoding. To address the limitation, we propose to transform it to a object coordinate system defined by the rotated bounding box.  As illustrated in~\Cref{fig:rotatedRPE}, an offset vector in the world coordinate system can be transformed to the object coordinate system $(x_\theta,y_\theta)$ following:
\begin{equation} \label{eq.rotate}
    \begin{bmatrix}
        \Delta x_\theta \\
        \Delta y_\theta \\
        \Delta z_\theta \\
    \end{bmatrix}
     =   \setlength{\arraycolsep}{2.0pt}
    \begin{bmatrix}
        \cos\theta & -\sin\theta & 0 \\
        \sin\theta & \cos\theta & 0 \\
        0 & 0 & 1
    \end{bmatrix}^T
    \begin{bmatrix}
        \Delta x \\
        \Delta y \\
        \Delta z \\
    \end{bmatrix}
    =
    \mathbf{R}_\theta^T \Delta\mathbf{p},
\end{equation}
where $\Delta\mathbf{p}$ is an element of ${\Delta\mathbf{P}_{i}}$. We use the other transformations in Equation~\ref{eq.rpe_transform} and Equation~\ref{eq.rpe_add} to get the final normalized relative position bias item that models the rotated 3D bounding box position information. We perform 3DV-RPE operations for different Transformer decoder layers by default.

\paragraph{Efficient implementation.} 
A naive implementation has high GPU memory consumption due to the large number of combinations between the object queries (each object query predicts a 3D bounding box) and the key-value points (output by the encoder), i.e. $\mathsf{K}{\times}\mathsf{N}{=}1,024\times4,096$, which makes it hard to train and deploy.

To solve this challenge, we use a smaller pre-defined 3DV-RPE table of shape: $\mathbf{T}\in \mathbb{R}^{10\times10\times10}$. We apply the non-linear projection $\mathcal{F}$ on this 3DV-RPE table and do volumetric (5-D) $\operatorname{grid\_sample}$ on the transformed 3DV-RPE table as follows:
\begin{equation}
\begin{aligned}
\label{eq.rpe_mlp_4}
{\mathbf{P}_i} = \operatorname{grid\_sample}(\operatorname{MLP}_{i}(\mathcal{F}(\mathbf{T})),\; {\Delta\mathbf{P}_{i}}).
\end{aligned}
\end{equation}

\vspace{3mm}
\subsection{DETR with 3DV-RPE}
\noindent\textbf{Framework.} We extend the original plain Transformer decoder, which consists of a stack of decoder layers and was designed for 2D object detection, to detect 3D bounding boxes from the irregular 3D points. Our approach has two steps: (i) as the first decoder layer has no access to coarse 3D bounding boxes, we employ a light-weight FFN to predict the initial 3D bounding boxes and feed the top confident ones to the first Transformer decoder layer (e.g., $\{\theta^0, x^0, y^0, z^0, w^0, l^0, h^0\}$); and (ii) we update the bounding box predictions with the output of each Transformer decoder layer and use them to compute the modulation term in the multi-head cross-attention.

Figure~\ref{fig:pipeline_details} illustrates more details of the DETR with our 3DV-RPE. For instance, we employ only the 3D content query $\mathbf{Q}_c$ as the input for the first decoder layer and use the decoder output embeddings $\mathbf{Q}^{i-1}$ from the $(i-1)$-th decoder layer as the input for the $i$-th decoder layer. We also apply MLP projects to compute the absolute position encodings of the 3D bounding boxes by default. We set the number of decoder layers as $8$ following~\cite{misra2021-3detr}.
We predict the 3D bounding box delta target based on the initial prediction such as $\{\theta^0, x^0, y^0, z^0, w^0, l^0, h^0\}$ in all the Transformer decoder layers.

\vspace{1mm}
\noindent\textbf{Visualization.}
Figure~\ref{fig:attention_maps_rpe} shows the relative position attention maps learned with the 3DV-RPE scheme. We show the attention maps for $8$ vertices in the first $4$ columns and the merged ones in the last column. The visualization results show that (i) our 3DV-RPE can enhance the inner 3D box regions relative to each vertex position and (ii) combining the eight relative position attention maps can accurately localize the regions within the bounding box. We also show that 3DV-RPE can localize the extremity positions on the 3D object surface in the experiments.

\section{Experiment}
\subsection{Datasets and metrics}
\noindent \textbf{Datasets.} We evaluate our approach on two challenging 3D indoor object detection benchmarks including:

\noindent \emph{ScanNetV2}~\cite{dai2017scannet}: ScanNetV2 consists of 3D meshes recovered from RGB-D videos captured in various indoor scenes. It has about $12$K training meshes and $312$ validation meshes, each annotated with semantic and instance segmentation masks for around $18$ classes of objects. We follow~\cite{qi2019deep} to extract the point clouds from the meshes.

\noindent \emph{SUN RGB-D}~\cite{song2015sun}: SUN RGB-D is a single-view RGB-D image dataset. It has about $5$K images for both training and validation sets. Each image is annotated with oriented 3D bounding boxes for $37$ classes of objects. We follow VoteNet~\cite{qi2019deep} to convert the RGB-D image to the point clouds using the camera parameters and evaluate our approach on the $10$ most common classes of objects.

\vspace{1mm}
\noindent \textbf{Metrics.} We report the standard mean Average Precision (mAP) under different IoU thresholds, \ie AP$_{25}$ for $0.25$ IoU threshold and AP$_{50}$ for $0.5$ IoU threshold. 

\renewcommand{\arraystretch}{1.45}
\begin{table}[!t]
\centering
\footnotesize
\setlength{\tabcolsep}{12pt}
\resizebox{1.0\linewidth}{!}
{
\begin{tabular}{@{}l|cccc}
\multirow{2}{*}{Method} & \multicolumn{2}{c}{ScanNetV2} & \multicolumn{2}{c}{SUN RGB-D} \\ 
& AP$_{25}$ & AP$_{50}$ & AP$_{25}$ & AP$_{50}$ \\
\shline
VoteNet~\cite{qi2019deep} & $58.6$ & $33.5$ & $57.7$ & - \\

HGNet~\cite{chen20hgnet} & $61.3$ & $34.4$ & $61.6$ & - \\

3D-MPA~\cite{engelmann203dmpa} & $64.2$ & $49.2$ & - & - \\

MLCVNet~\cite{xie2020mlcvnet} & $64.5$ & $41.4$ & $59.8$ & - \\

GSDN~\cite{gwak2020generative} & $62.8$ & $34.8$ & - & - \\

H3DNet~\cite{zhang2020h3dnet} & $67.2$ & $48.1$ & $60.1$ & $39.0$ \\

BRNet~\cite{cheng21brnet} & $66.1$ & $50.9$ & $61.1$ & $43.7$ \\

3DETR~\cite{misra2021-3detr} & $65.0$ & $47.0$ & $59.1$ & $32.7$ \\

VENet~\cite{xie21venet} & $67.7$ & - & $62.5$ & $39.2$ \\

Group-Free~\cite{liu2021group} & $69.1$ & $52.8$ & $63.0$ & $45.2$ \\

RBGNet~\cite{wang22rbgnet} & $70.6$ & $55.2$ & $64.1$ & $47.2$ \\

HyperDet3D~\cite{zheng22hyperdet3d} & $70.9$ & $57.2$ & $63.5$ & $47.3$ \\

FCAF3D~\cite{rukhovich2022fcaf3d} & $71.5$ & $57.3$ & $64.2$ & $48.9$ \\

TR3D~\cite{rukhovich23tr3d} & $72.9$ & $59.3$ & $67.1$ & $50.4$ \\

CAGroup3D~\cite{wang2022cagroup3d} & $75.1$ & $61.3$ & $66.8$ & $50.2$ \\
\rowcolor{gray!10}V-DETR & ${77.4}$ & ${65.0}$ & ${67.5}$ & ${50.4}$ \\
\rowcolor{gray!10}V-DETR (TTA) & $\bf{77.8}$ & $\bf{66.0}$ & $\bf{68.0}$ & $\bf{51.1}$ \\

\shline
\multicolumn{5}{c}{\emph{Average Results under $25\times$ trials}}     \\
\hline
Group-Free~\cite{liu2021group}  & $68.6$ & $51.8$ & $62.6$ & $44.4$ \\
RBGNet~\cite{wang22rbgnet} & $69.9$ & $54.7$ & $63.6$ & $46.3$ \\
FCAF3D~\cite{rukhovich2022fcaf3d}  & $70.7$ & $56.0$ & $63.8$ & $48.2$ \\
TR3D~\cite{rukhovich23tr3d} & $72.0$ & $57.4$ & $66.3$ & $49.6$ \\
CAGroup3D~\cite{wang2022cagroup3d} & $74.5$ & $60.3$ & $66.4$ & $49.5$ \\
\rowcolor{gray!10}V-DETR  & ${76.8}$ & ${64.5}$ & ${66.8}$ & ${49.7}$ \\
\rowcolor{gray!10}V-DETR (TTA) & $\bf{77.0}$ & $\bf{65.3}$ & $\bf{67.5}$ & $\bf{50.0}$ \\
\end{tabular}
}
\caption{\small{System-level comparison with the state-of-the-art on ScanNetV2 and SUN RGB-D. TTA: test-time augmentation.}
}
\vspace{-5mm}
\label{tab:sota_comparison}
\end{table}

\subsection{Implementation details}

\vspace{1mm}
\noindent \textbf{Training.} We use the AdamW optimizer~\cite{Loshchilov2019adamw} with the base learning rate $7$e-$4$, the batch size $8$, and the weight decay $0.1$. The learning rate is warmed up for $9$ epochs, then is dropped to $1$e-$6$ using the cosine schedule during the entire training process.
We use gradient clipping to stabilize the training. We train for $360$ epochs on ScanNetV2 and $240$ epochs on SUN RGB-D in all experiments except for the system-level comparisons, where we train for $540$ epochs on ScanNetV2.
We use the standard data augmentations including random cropping (at least $30$K points), random sampling ($100$K points), random flipping (p=0.5), random rotation along the z-axis (-$5^{\circ}$, $5^{\circ}$), random translation (-$0.4$, $0.4$), random scaling ($0.6$, $1.4$).
We also use the one-to-many matching~\cite{jia2022detrs} to speed up the convergence speed with more rich and informative positive samples.

\vspace{1mm}
\noindent \textbf{Inference.}
We process the entire point clouds of each scene and generate the bounding box proposals. We use 3D NMS to suppress the duplicated proposals in the one-to-many matching setting, which is not needed in the one-to-one matching setting. We also use test-time augmentation, i.e., flipping, by default unless specified otherwise.

\subsection{Comparisons with Previous Systems}
In Table~\ref{tab:sota_comparison}, we compare our method with the state-of-the-art methods from previous works at the system level.
These methods use different techniques, so we cannot compare them in a controlled way. According to the results, we show that our method performs the best either measured by the highest performance or the average results under multiple trials. For example, on ScanNetV2 \texttt{val} set, our method achieves AP$_{25}$=$77.8\%$ and AP$_{50}$=$66.0\%$, which surpasses the latest state-of-the-art CAGroup3D that reports AP$_{25}$=$75.1\%$ and AP$_{50}$=$61.3\%$.
Notably, on ScanNetV2, we observe more significant gains on AP$_{50}$ (+$4.7\%$) that requires more accurate localization, i.e., under a higher IoU threshold.
We also observe consistent gains on both AP$_{25}$ and AP$_{50}$ on SUN RGB-D.

\subsection{3DV-RPE Ablation Experiments}
We conduct all the following ablation experiments on ScanNetV2 except for the ablation experiments on the coordinate system, where we report the results on SUN RGB-D.
\vspace{1mm}

\begin{figure}[t]
\centering
\includegraphics[width=0.475\textwidth]{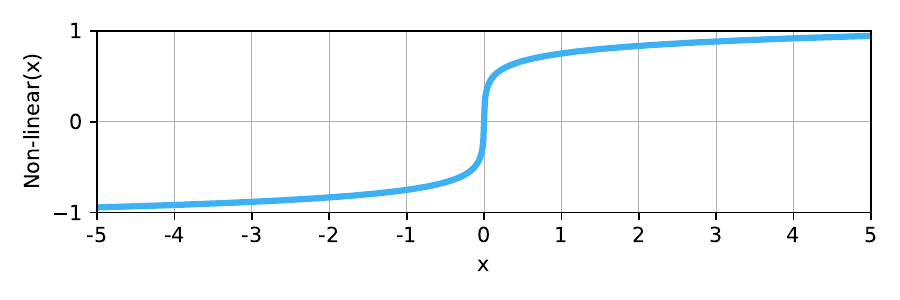}
\caption{\small{{Illustrating the curve of signed log transform function.}}}
\label{fig:curve_nonlinear}
\end{figure}

\begin{table}[t]
\footnotesize
\renewcommand{\arraystretch}{1.2}
\centering
\begin{minipage}{1\linewidth}
{\begin{center}
\tablestyle{20pt}{1.35}
\resizebox{1.0\linewidth}{!}
{
\begin{tabular}{l|cc}
  $\mathcal{F}(\cdot)$ & $\operatorname{AP}_{25}$ & $\operatorname{AP}_{50}$ \\
    \shline
    $\mathcal{F}(x)$ = $x$ &$69.6$ &$48.2$ \\
    $\mathcal{F}(x)$ = $x/(1+|x|)$ &$76.0$ &$62.6$ \\
    $\mathcal{F}(x)$ = $\operatorname{tanh}(x)$ &$76.3$ &$62.6$ \\
    $\mathcal{F}(x)$ = $x/\sqrt{1+x^2}$ &$76.6$ &$63.0$ \\
    \rowcolor{gray!10}$\mathcal{F}(x)$ = $\operatorname{sign}(x)\log(1+|x|)$ & $76.7$ & $65.0$ \\  
\end{tabular}
}
\end{center}}
\end{minipage}
\caption{\small{Effect of non-linear transform within 3DV-RPE}.}
\label{tab:3dv_rpe_nonlinear}
\end{table}

\begin{table}[t]
\begin{minipage}[t]{1\linewidth}
\vspace{2mm}
\centering
\setlength{\tabcolsep}{32pt}
\footnotesize
\renewcommand{\arraystretch}{1.35}
\resizebox{1.0\linewidth}{!}
{
\begin{tabular}{c|cc}
\# vertex  &$\operatorname{AP}_{25}$ & $\operatorname{AP}_{50}$ \\
\shline
$1$ & $73.4$ & $54.8$     \\
$2$ & $76.1$ & $63.1$     \\
$4$ & $76.3$ & $63.4$     \\
\rowcolor{gray!10}$8$ & $76.7$ & $65.0$     \\
\end{tabular}
}
\caption{\small{{
Effect of the number of vertex within 3DV-RPE.}}
}
\label{tab:3dv_rpe_num_vertex}
\end{minipage}
\end{table}

\noindent \textbf{Non-linear transform.}
Table~\ref{tab:3dv_rpe_nonlinear} shows the effect of different non-linear transform functions. The results show that the signed log function performs the best. Figure~\ref{fig:curve_nonlinear} illustrates the curve of the signed log function and shows how it magnifies small changes in smaller ranges.
Therefore, we choose the signed log function by default.

\vspace{1mm}
\noindent \textbf{Number of vertex.}
Table~\ref{tab:3dv_rpe_num_vertex} shows the effect of different numbers of vertices for computing the relative position bias term. The results show that using $8$ vertices performs the best, so we use this setting by default.
We attribute their close performances to the fact that they essentially share the same minimal and maximal XYZ values when using fewer vertices such as $2$ or $4$, which is caused by the zero rotation angles on ScanNetV2.

\vspace{1mm}
\noindent \textbf{Coordinate system on SUN RGB-D.}
We evaluate the effect of the coordinate system on calculating the relative positions in our 3DV-RPE on SUN RGB-D, which requires predicting the rotation angle along the $z$-axis. Table~\ref{tab:3dv_rpe_rotation} shows the results. We find that transforming the relative offsets from the world coordinate system to the object coordinate system significantly improves the performance, e.g., $\operatorname{AP}_{25}$ and $\operatorname{AP}_{50}$ increase by +$2.2\%$ and $4.2\%$, respectively.

\vspace{1mm}
\noindent \textbf{Comparison with 3D box mask.}
Table~\ref{tab:effect_cross_attn} compares our 3DV-RPE with a 3D box mask method, which sets the relative position bias term to $-\infty$ for positions outside the 3D bounding box and $0$ otherwise. The results show that (i) the 3D box mask method achieves strong results on $\operatorname{AP}_{25}$, and (ii) our 3DV-RPE significantly improves over the 3D box mask method on $\operatorname{AP}_{50}$. We speculate that our 3DV-RPE performs better because the 3D box mask method suffers from error accumulation from the previous decoder layers and cannot be optimized end-to-end.
We also report the results of combining the 3D box mask and 3DV-RPE, which performs better than the 3D box mask scheme but worse than our 3DV-RPE. This verifies that our 3DV-RPE can learn to (i) exploit more accurate geometric structure information within the 3D bounding box and (ii) benefit from capturing useful long-range context information outside the box. Moreover, we report the results with longer training epochs and observe that the gap between the 3D box mask and 3DV-RPE remains, thus further demonstrating the advantages of our approach.

\begin{table}[t]
\footnotesize
\renewcommand{\arraystretch}{1.1}
\centering
\begin{minipage}{1\linewidth}
{\begin{center}
\tablestyle{24pt}{1.3}
\resizebox{1.0\linewidth}{!}
{
\begin{tabular}{c|cc}
coordinate system  & $\operatorname{AP}_{25}$ & $\operatorname{AP}_{50}$ \\
    \shline
   world coord.& $65.8$ & $46.9$ \\ 
   \rowcolor{gray!10}object coord.& $68.0$ & $51.1$ \\ 
\end{tabular}
}
\end{center}}
\end{minipage}
\caption{\small Effect of the coordinate system on SUN RGB-D.}
\label{tab:3dv_rpe_rotation}
\end{table}

\begin{table}[t]
\footnotesize
\renewcommand{\arraystretch}{1.2}
\centering
\begin{minipage}{1\linewidth}{\begin{center}
\tablestyle{5pt}{1.3}
\setlength{\tabcolsep}{12pt}
\resizebox{1.0\linewidth}{!}
{
\begin{tabular}{l|ccc}
   attention modulation & \#epochs &  $\operatorname{AP}_{25}$ & $\operatorname{AP}_{50}$ \\
    \shline
    None & $360$ & $68.8$ & $44.5$ \\ 
    3D box mask  & $360$ & $74.0$ & $59.1$ \\ 
    \rowcolor{gray!10}3DV-RPE  & $360$ & $76.7$ & $65.0$ \\
    3D box mask + 3DV-RPE  & $360$ & $76.0$ & $62.7$\\ \hline
    None  & $540$ & $71.4$ & $47.6$ \\ 
    3D box mask   & $540$ & $75.1$ & $60.8$ \\ 
    \rowcolor{gray!10}3DV-RPE  & $540$ & ${77.8}$ & $66.0$\\
    3D box mask + 3DV-RPE  & $540$ &$77.0$ & $63.5$\\ 
\end{tabular}
}
\end{center}}
\end{minipage}
\caption{\small Effect of the attention modulation choices.}
\label{tab:effect_cross_attn}
\end{table}

\subsection{Other Ablation Experiments}
We study the effect of the other components in the following experiments.

\vspace{1mm}
\noindent \textbf{Encoder choice.}
Table~\ref{tab:effect_backbone} compares the results of using different encoder architectures. We find that using a sparse 3D version of ResNet$34$ with an FPN neck achieves the best results. Therefore, we use ResNet$34$ + FPN as our default encoder.
\begin{table}[t]
\footnotesize
\renewcommand{\arraystretch}{1.2}
\centering
\begin{minipage}{1\linewidth}
{\begin{center}
\tablestyle{24pt}{1.3}
\resizebox{1.0\linewidth}{!}
{
\begin{tabular}{l|cc}
   encoder  & $\operatorname{AP}_{25}$ & $\operatorname{AP}_{50}$ \\
    \shline
   PointNet + Tran.Enc. & $73.6$ & $60.1$ \\ 
   \rowcolor{gray!10}ResNet$34$ + FPN & $76.7$ & $65.0$ \\ 
\end{tabular}
}
\end{center}}
\end{minipage}
\caption{\small Effect of the encoder choice.}
\label{tab:effect_backbone}
\end{table}

\begin{table}[t]
\footnotesize
\renewcommand{\arraystretch}{1.2}
\centering
\begin{minipage}{1\linewidth}
{\begin{center}
\tablestyle{25pt}{1.3}
\resizebox{1.0\linewidth}{!}
{
\begin{tabular}{c|cc}
  object-normalize  & $\operatorname{AP}_{25}$ & $\operatorname{AP}_{50}$ \\
    \shline
  \xmark &$74.9$& $61.1$ \\ 
  \rowcolor{gray!10}\cmark & $76.7$ & $65.0$ \\ 
\end{tabular}
}
\end{center}}
\end{minipage}
\caption{\small Effect of the object-normalized box parameterization.}
\label{tab:effect_reparam}
\end{table}

\begin{table}[t]
\begin{minipage}[t]{1\linewidth}
\vspace{2mm}
\centering
\setlength{\tabcolsep}{15pt}
\footnotesize
\renewcommand{\arraystretch}{1.2}
\resizebox{1.0\linewidth}{!}
{
\begin{tabular}{l|c|cc}
method & voxel expansion  &$\operatorname{AP}_{25}$ & $\operatorname{AP}_{50}$ \\
\shline
\multirow{2}{*}{FCAF3D} & \xmark & $67.5$ & $52.4$         \\
& \cmark & $70.5$ & $54.8$ \\ \hline
\multirow{2}{*}{Ours} & \cellcolor{gray!10}\xmark & \cellcolor{gray!10}$76.7$ & \cellcolor{gray!10}$65.0$         \\
& \cmark & $75.5$ & $62.0$        \\
\end{tabular}
}
\caption{\small{{
Effect of voxel expansion.}}
}
\label{tab:voxel_expan}
\end{minipage}
\end{table}

\begin{figure*}[t]
\centering
\includegraphics[height=0.28\columnwidth]{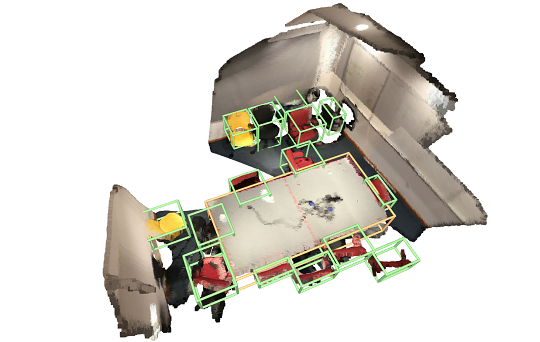}
\hfill
\includegraphics[height=0.28\columnwidth]{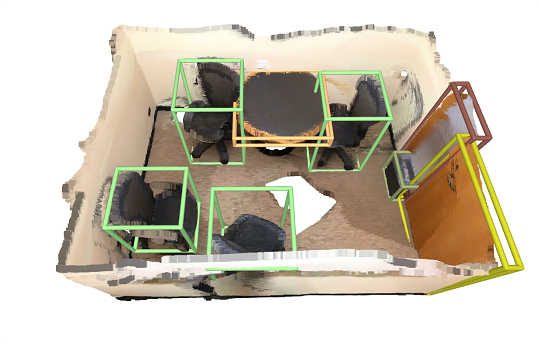}
\hfill
\includegraphics[height=0.28\columnwidth]{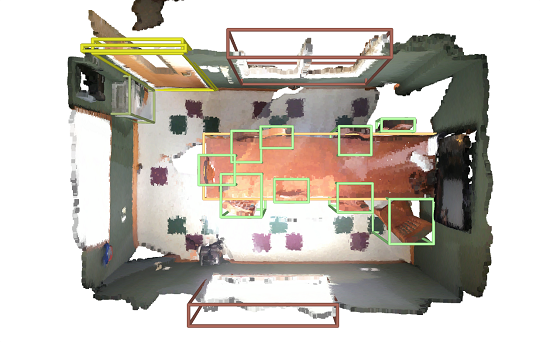}
\hfill
\includegraphics[height=0.28\columnwidth]{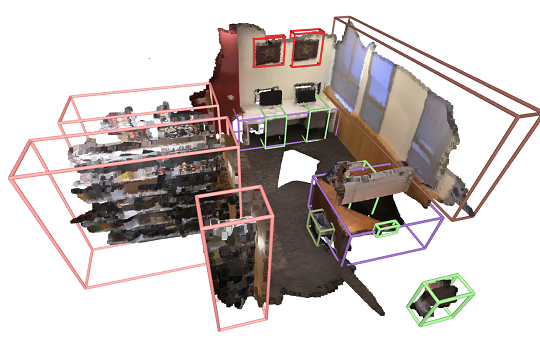} \\
\includegraphics[height=0.28\columnwidth]{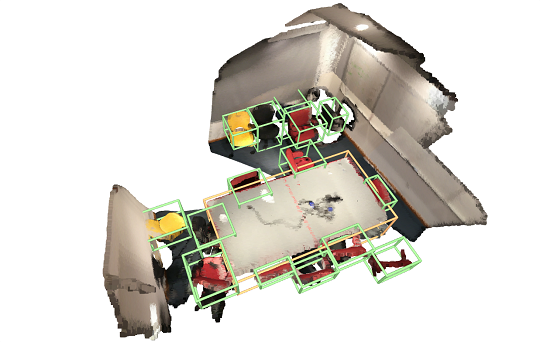}
\hfill
\includegraphics[height=0.28\columnwidth]{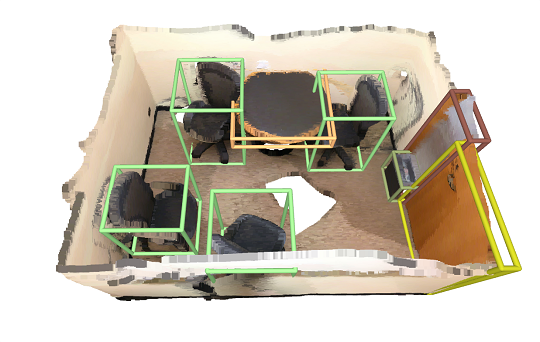}
\hfill
\includegraphics[height=0.28\columnwidth]{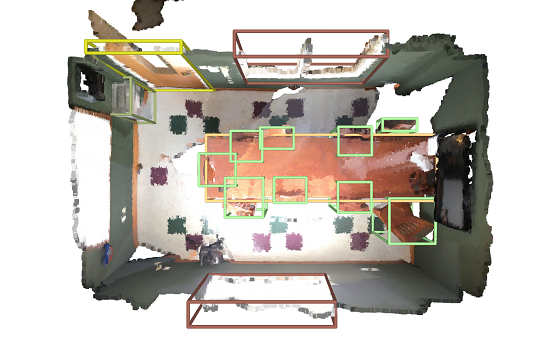}
\hfill
\includegraphics[height=0.28\columnwidth]{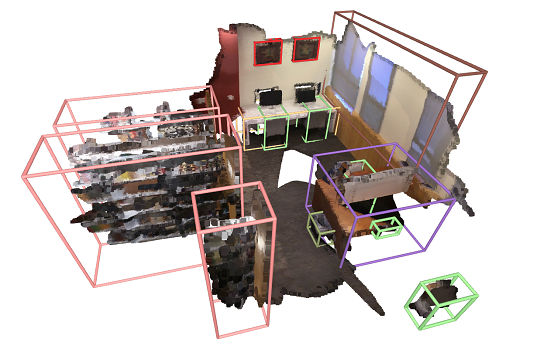}

\caption{\small{
Qualitative results of 3D object detection on ScanNetV2. The ground-truth is shown in the first row and our method's detection results are shown in the second row.
}}
\label{fig:qualitative}
\end{figure*}

\vspace{1mm}
\noindent \textbf{Object-normalized box parameterization.}
In Table~\ref{tab:effect_reparam}, we show the effect of using object-normalized box parameterization.
We find using the object-normalized scheme significantly boosts the $\operatorname{AP}_{50}$ from $
61.1$ to $65.0$.

\vspace{1mm}
\noindent \textbf{Voxel expansion.}
Table~\ref{tab:voxel_expan} evaluates the effect of using voxel expansion in the FPN neck when the encoder is ResNet$34$ + FPN. We also compare our results with the recent FCAF3D method. The results show that (i) voxel expansion is crucial for FCAF3D, which relies on building virtual center features; and (ii) voxel expansion degrades the performance when using DETR, which might lose the original accurate 3D surface information. Therefore, we demonstrate an important advantage of using DETR-based approaches, i.e., they do not require complicated voxel expansion operations.

\begin{figure}[t]
\centering
\includegraphics[height=0.23\columnwidth, trim={50 0 70 0},clip]{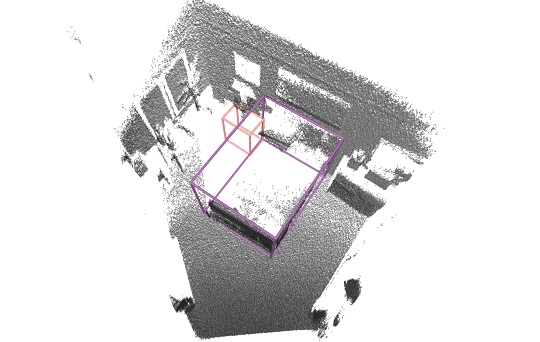}
\includegraphics[height=0.23\columnwidth, trim={50 0 50 0},clip]{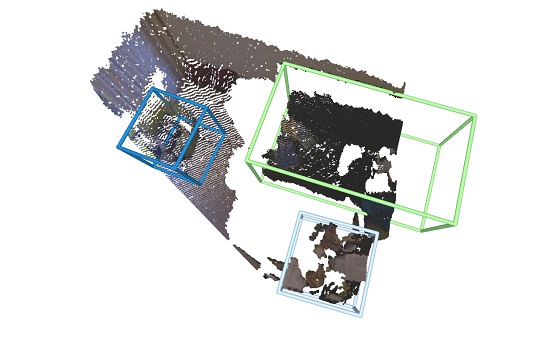}
\includegraphics[height=0.23\columnwidth, trim={50 0 50 0},clip]{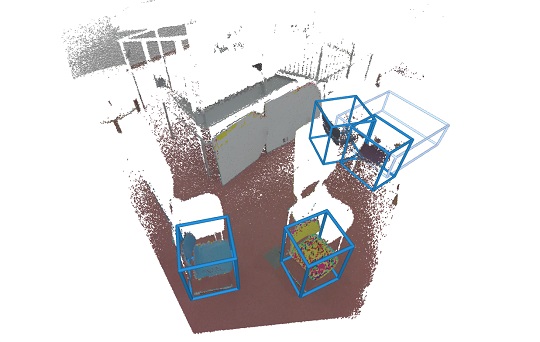}
\hfill \\
\includegraphics[height=0.23\columnwidth,trim={50 0 70 0},clip]{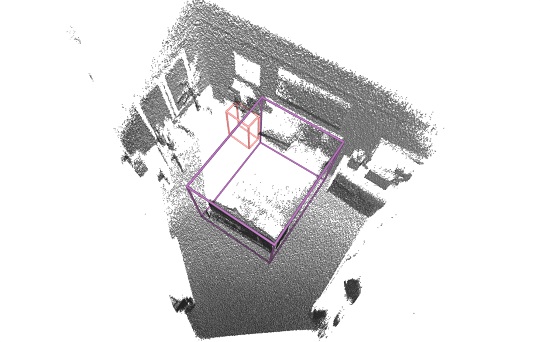}
\includegraphics[height=0.23\columnwidth,trim={50 0 50 0},clip]{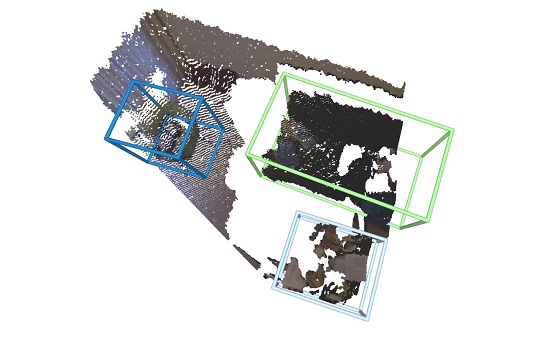}
\includegraphics[height=0.23\columnwidth,trim={50 0 50 0},clip]{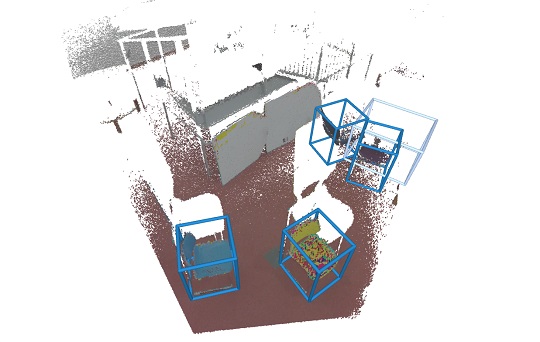}
\caption{\small{
Qualitative results of 3D object detection on SUN RGB-D. The ground-truth is shown in the first row and our method's detection results are shown in the second row.
}}
\label{fig:qualitative_sunrgbd}
\end{figure}

\vspace{1mm}
\noindent \textbf{Qualitative comparisons.} 
We show some examples of V-DETR detection results on ScanNet in Figure~\ref{fig:qualitative}, where the scenes are diverse and challenging with clutter, partiality, scanning artifacts, etc.
Our V-DETR performs well despite these challenges. For example, it detects most of the chairs in the scene shown in the $1$-st column.
Figure~\ref{fig:qualitative_sunrgbd} shows some examples of our prediction results on SUN RGB-D. Accordingly, we find our V-DETR can handle the rotated bounding boxes under various challenging rotation angles.
We include more qualitative comparisons in the supplementary material.

\noindent \emph{More ablation experiments.} 
We provide more ablation studies on the effects of using different shapes for the pre-defined 3DV-RPE table, one-to-many matching, the number of points in training and testing, and other factors in the supplementary material.

\section{Conclusion}
In this work, we have shown how to make DETR-based approaches competitive for indoor 3D object detection tasks. The key contribution is an effective 3D vertex relative position encoding (3DV-RPE) scheme that can model the accurate position information in the irregular sparse 3D point cloud directly. We demonstrate the advantages of our approach by achieving strong results on two challenging 3D detection benchmarks. We also plan to extend our approach to outdoor 3D object detection tasks, which differ from most existing methods that rely on modern 2D DETR-based detectors by converting 3D points to a 2D bird-eye-view plane.
We hope our approach can show the potential for unifying the object detection architecture design for indoor and outdoor 3D detection tasks.

{\small
\bibliographystyle{ieee}
\bibliography{egbib}

\begin{thebibliography}{10}\itemsep=-1pt

\bibitem{carion2020end}
N.~Carion, F.~Massa, G.~Synnaeve, N.~Usunier, A.~Kirillov, and S.~Zagoruyko.
\newblock End-to-end object detection with transformers.
\newblock In {\em ECCV}, pages 213--229, 2020.

\bibitem{chen20hgnet}
J.~Chen, B.~Lei, Q.~Song, H.~Ying, D.~Z. Chen, and J.~Wu.
\newblock A hierarchical graph network for 3d object detection on point clouds.
\newblock In {\em CVPR}, pages 389--398, 2020.

\bibitem{chen2022group}
Q.~Chen, X.~Chen, G.~Zeng, and J.~Wang.
\newblock Group detr: Fast training convergence with decoupled one-to-many
  label assignment.
\newblock {\em arXiv preprint arXiv:2207.13085}, 2022.

\bibitem{cheng21brnet}
B.~Cheng, L.~Sheng, S.~Shi, M.~Yang, and D.~Xu.
\newblock Back-tracing representative points for voting-based 3d object
  detection in point clouds.
\newblock In {\em CVPR}, pages 8963--8972.

\bibitem{dai2017scannet}
A.~Dai, A.~X. Chang, M.~Savva, M.~Halber, T.~Funkhouser, and M.~Nie{\ss}ner.
\newblock Scannet: Richly-annotated 3d reconstructions of indoor scenes.
\newblock In {\em CVPR}, pages 5828--5839, 2017.

\bibitem{dai2021dynamic}
X.~Dai, Y.~Chen, J.~Yang, P.~Zhang, L.~Yuan, and L.~Zhang.
\newblock Dynamic detr: End-to-end object detection with dynamic attention.
\newblock In {\em ICCV}, pages 2988--2997, 2021.

\bibitem{engelmann203dmpa}
F.~Engelmann, M.~Bokeloh, A.~Fathi, B.~Leibe, and M.~Nie{\ss}ner.
\newblock 3d-mpa: Multi-proposal aggregation for 3d semantic instance
  segmentation.
\newblock In {\em CVPR}, pages 9028--9037.

\bibitem{fan2023super}
L.~Fan, Y.~Yang, F.~Wang, N.~Wang, and Z.~Zhang.
\newblock Super sparse 3d object detection.
\newblock {\em arXiv preprint arXiv:2301.02562}, 2023.

\bibitem{gao2021fast}
P.~Gao, M.~Zheng, X.~Wang, J.~Dai, and H.~Li.
\newblock Fast convergence of detr with spatially modulated co-attention.
\newblock In {\em ICCV}, pages 3621--3630, 2021.

\bibitem{girshick2015fast}
R.~Girshick.
\newblock Fast r-cnn.
\newblock In {\em ICCV}, pages 1440--1448, 2015.

\bibitem{gwak2020generative}
J.~Gwak, C.~Choy, and S.~Savarese.
\newblock Generative sparse detection networks for 3d single-shot object
  detection.
\newblock In {\em ECCV}, pages 297--313. Springer, 2020.

\bibitem{jia2022detrs}
D.~Jia, Y.~Yuan, H.~He, X.~Wu, H.~Yu, W.~Lin, L.~Sun, C.~Zhang, and H.~Hu.
\newblock Detrs with hybrid matching.
\newblock {\em arXiv preprint arXiv:2207.13080}, 2022.

\bibitem{lai2022stratified}
X.~Lai, J.~Liu, L.~Jiang, L.~Wang, H.~Zhao, S.~Liu, X.~Qi, and J.~Jia.
\newblock Stratified transformer for 3d point cloud segmentation.
\newblock In {\em arXiv preprint arXiv:2203.14508}, 2022.

\bibitem{lang2019pointpillars}
A.~H. Lang, S.~Vora, H.~Caesar, L.~Zhou, J.~Yang, and O.~Beijbom.
\newblock Pointpillars: Fast encoders for object detection from point clouds.
\newblock In {\em CVPR}, pages 12697--12705, 2019.

\bibitem{li2022dn}
F.~Li, H.~Zhang, S.~Liu, J.~Guo, L.~M. Ni, and L.~Zhang.
\newblock Dn-detr: Accelerate detr training by introducing query denoising.
\newblock {\em arXiv preprint arXiv:2203.01305}, 2022.

\bibitem{lin2017focal}
T.-Y. Lin, P.~Goyal, R.~Girshick, K.~He, and P.~Doll{\'a}r.
\newblock Focal loss for dense object detection.
\newblock In {\em ICCV}, pages 2980--2988, 2017.

\bibitem{liu2022dab}
S.~Liu, F.~Li, H.~Zhang, X.~Yang, X.~Qi, H.~Su, J.~Zhu, and L.~Zhang.
\newblock Dab-detr: Dynamic anchor boxes are better queries for detr.
\newblock {\em arXiv preprint arXiv:2201.12329}, 2022.

\bibitem{liu2016ssd}
W.~Liu, D.~Anguelov, D.~Erhan, C.~Szegedy, S.~Reed, C.-Y. Fu, and A.~C. Berg.
\newblock Ssd: Single shot multibox detector.
\newblock In {\em ECCV}, pages 21--37. Springer, 2016.

\bibitem{liu2021swin}
Z.~Liu, Y.~Lin, Y.~Cao, H.~Hu, Y.~Wei, Z.~Zhang, S.~Lin, and B.~Guo.
\newblock Swin transformer: Hierarchical vision transformer using shifted
  windows.
\newblock In {\em CVPR}, pages 10012--10022, 2021.

\bibitem{liu2021group}
Z.~Liu, Z.~Zhang, Y.~Cao, H.~Hu, and X.~Tong.
\newblock Group-free 3d object detection via transformers.
\newblock In {\em CVPR}, pages 2949--2958, 2021.

\bibitem{Loshchilov2019adamw}
I.~Loshchilov and F.~Hutter.
\newblock Decoupled weight decay regularization.
\newblock In {\em International Conference on Learning Representations}, 2019.

\bibitem{meng2021CondDETR}
D.~Meng, X.~Chen, Z.~Fan, G.~Zeng, H.~Li, Y.~Yuan, L.~Sun, and J.~Wang.
\newblock Conditional detr for fast training convergence.
\newblock In {\em ICCV}, 2021.

\bibitem{misra2021-3detr}
I.~Misra, R.~Girdhar, and A.~Joulin.
\newblock {An End-to-End Transformer Model for 3D Object Detection}.
\newblock In {\em {ICCV}}, 2021.

\bibitem{neubeck2006efficient}
A.~Neubeck and L.~Van~Gool.
\newblock Efficient non-maximum suppression.
\newblock In {\em ICPR}, volume~3, pages 850--855, 2006.

\bibitem{qi2019deep}
C.~R. Qi, O.~Litany, K.~He, and L.~J. Guibas.
\newblock Deep hough voting for 3d object detection in point clouds.
\newblock In {\em ICCV}, pages 9276--9285, 2019.

\bibitem{ren2015faster}
S.~Ren, K.~He, R.~Girshick, and J.~Sun.
\newblock Faster r-cnn: Towards real-time object detection with region proposal
  networks.
\newblock {\em Advances in neural information processing systems}, 28, 2015.

\bibitem{rukhovich2022fcaf3d}
D.~Rukhovich, A.~Vorontsova, and A.~Konushin.
\newblock Fcaf3d: fully convolutional anchor-free 3d object detection.
\newblock In {\em ECCV}, pages 477--493. Springer, 2022.

\bibitem{rukhovich23tr3d}
D.~Rukhovich, A.~Vorontsova, and A.~Konushin.
\newblock {TR3D:} towards real-time indoor 3d object detection.
\newblock {\em CoRR}, abs/2302.02858, 2023.

\bibitem{song2015sun}
S.~Song, S.~P. Lichtenberg, and J.~Xiao.
\newblock Sun rgb-d: A rgb-d scene understanding benchmark suite.
\newblock In {\em CVPR}, pages 567--576, 2015.

\bibitem{Vaswani2017attention}
A.~Vaswani, N.~Shazeer, N.~Parmar, J.~Uszkoreit, L.~Jones, A.~N. Gomez,
  L.~Kaiser, and I.~Polosukhin.
\newblock Attention is all you need.
\newblock In {\em NeurIPS}, pages 5998--6008, 2017.

\bibitem{wang2022cagroup3d}
H.~Wang, L.~Ding, S.~Dong, S.~Shi, A.~Li, J.~Li, Z.~Li, and L.~Wang.
\newblock Cagroup3d: Class-aware grouping for 3d object detection on point
  clouds.
\newblock {\em arXiv preprint arXiv:2210.04264}, 2022.

\bibitem{wang22rbgnet}
H.~Wang, S.~Shi, Z.~Yang, R.~Fang, Q.~Qian, H.~Li, B.~Schiele, and L.~Wang.
\newblock Rbgnet: Ray-based grouping for 3d object detection.
\newblock In {\em CVPR}, pages 1100--1109.

\bibitem{wang2021anchor}
Y.~Wang, X.~Zhang, T.~Yang, and J.~Sun.
\newblock Anchor detr: Query design for transformer-based detector, 2021.

\bibitem{xie21venet}
Q.~Xie, Y.~Lai, J.~Wu, Z.~Wang, D.~Lu, M.~Wei, and J.~Wang.
\newblock Venet: Voting enhancement network for 3d object detection.
\newblock In {\em ICCV}, pages 3692--3701.

\bibitem{xie2020mlcvnet}
Q.~Xie, Y.-K. Lai, J.~Wu, Z.~Wang, Y.~Zhang, K.~Xu, and J.~Wang.
\newblock Mlcvnet: Multi-level context votenet for 3d object detection.
\newblock In {\em CVPR}, pages 10447--10456, 2020.

\bibitem{yan2018second}
Y.~Yan, Y.~Mao, and B.~Li.
\newblock Second: Sparsely embedded convolutional detection.
\newblock {\em Sensors}, 18(10):3337, 2018.

\bibitem{yang2022unified}
Z.~Yang, L.~Jiang, Y.~Sun, B.~Schiele, and J.~Jia.
\newblock A unified query-based paradigm for point cloud understanding.
\newblock In {\em CVPR}, pages 8541--8551, 2022.

\bibitem{yin2021center}
T.~Yin, X.~Zhou, and P.~Krahenbuhl.
\newblock Center-based 3d object detection and tracking.
\newblock In {\em CVPR}, pages 11784--11793, 2021.

\bibitem{zhang2022dino}
H.~Zhang, F.~Li, S.~Liu, L.~Zhang, H.~Su, J.~Zhu, L.~M. Ni, and H.-Y. Shum.
\newblock Dino: Detr with improved denoising anchor boxes for end-to-end object
  detection.
\newblock {\em arXiv preprint arXiv:2203.03605}, 2022.

\bibitem{zhang2020h3dnet}
Z.~Zhang, B.~Sun, H.~Yang, and Q.~Huang.
\newblock H3dnet: 3d object detection using hybrid geometric primitives.
\newblock In {\em ECCV}, pages 311--329, 2020.

\bibitem{zheng22hyperdet3d}
Y.~Zheng, Y.~Duan, J.~Lu, J.~Zhou, and Q.~Tian.
\newblock Hyperdet3d: Learning a scene-conditioned 3d object detector.
\newblock In {\em CVPR}, pages 5575--5584.

\bibitem{zhou2018voxelnet}
Y.~Zhou and O.~Tuzel.
\newblock Voxelnet: End-to-end learning for point cloud based 3d object
  detection.
\newblock In {\em CVPR}, pages 4490--4499, 2018.

\bibitem{zhu2020deformable}
X.~Zhu, W.~Su, L.~Lu, B.~Li, X.~Wang, and J.~Dai.
\newblock Deformable detr: Deformable transformers for end-to-end object
  detection.
\newblock {\em arXiv preprint arXiv:2010.04159}, 2020.

\end{thebibliography}
}

\clearpage
\section{Supplementary}

\section*{A. More Ablation Experiments and Analysis}

\begin{table}[t]
\footnotesize
\renewcommand{\arraystretch}{1.2}
\centering
\begin{minipage}{1\linewidth}
{\begin{center}
\tablestyle{25pt}{1.2}
\resizebox{1.0\linewidth}{!}
{
\begin{tabular}{c|cc}
  Light-weight FFN  & $\operatorname{AP}_{25}$ & $\operatorname{AP}_{50}$ \\
    \shline
  \xmark & $76.6$ & $62.8$ \\ 
  \rowcolor{gray!10}\cmark & $76.7$ & $65.0$      \\ 
\end{tabular}
}
\end{center}}
\end{minipage}
\caption{\small Effect of light-weight FFN.}
\label{tab:ablate_ffn}
\end{table}

\begin{table}[t]
\footnotesize
\renewcommand{\arraystretch}{1.2}
\centering
\begin{minipage}{1\linewidth}
{\begin{center}
\tablestyle{30pt}{1.2}
\resizebox{1.0\linewidth}{!}
{
\begin{tabular}{c|cc}
  \# of points  & $\operatorname{AP}_{25}$ & $\operatorname{AP}_{50}$ \\
    \shline
  $20$K & $73.9$ & $61.6$ \\
  $40$K & $75.2$ & $62.4$ \\ 
  \rowcolor{gray!10}$100$K & $76.7$ & $65.0$ \\ 
\end{tabular}
}
\end{center}}
\end{minipage}
\caption{\small Effect of using more points during training and evaluation.}
\label{tab:ablate_point_num}
\end{table}

\begin{table}[t]
\footnotesize
\renewcommand{\arraystretch}{1.2}
\centering
\begin{minipage}{1\linewidth}
{\begin{center}
\tablestyle{10pt}{1.1}
\resizebox{1.0\linewidth}{!}
{
\begin{tabular}{ccc|cc}
   \# points & \#query & \#repeat number & $\operatorname{AP}_{25}$ & $\operatorname{AP}_{50}$ \\
    \shline
 \multirow{3}{*}{$40$K} &
   $256$ & $1$ & $74.3$ & $62.0$ \\
   & $512$ & $2$ & $75.6$ & $62.9$ \\
   & $1024$ & $4$ & $75.2$ & $62.4$
    \\ \hline
   \multirow{3}{*}{$100$K} & $256$ & $1$ & $75.3$ & $63.7$ \\
   & $512$ & $2$ & $76.4$ & $64.2$ \\
   & \cellcolor{gray!10}$1024$ & \cellcolor{gray!10}$4$ & \cellcolor{gray!10}$76.7$ & \cellcolor{gray!10}$65.0$          
\end{tabular}
}
\end{center}}
\end{minipage}
\caption{\small Effect of the one-to-many matching.}
\label{tab:effect_one2many}
\end{table}

\begin{table}[t]
\begin{minipage}[t]{1\linewidth}
\vspace{2mm}
\centering
\setlength{\tabcolsep}{23pt}
\footnotesize
\renewcommand{\arraystretch}{1.2}
\resizebox{1.0\linewidth}{!}
{
\begin{tabular}{c|cc}
3DV-RPE table shape  &$\operatorname{AP}_{25}$ & $\operatorname{AP}_{50}$ \\
\shline
$5\times5\times5$ & $76.7$ & $64.7$         \\
\rowcolor{gray!10}$10\times10\times10$ & $76.7$ & $65.0$        \\
$25\times25\times25$ & $76.7$ & $64.2$         \\
$50\times50\times50$ & $76.7$ & $64.3$        \\
\end{tabular}
}
\caption{\small{{
Effect of the pre-defined 3DV-RPE table shape.}}
}
\label{tab:ablate_shape}
\end{minipage}
\end{table}
\begin{table}[t]
\begin{minipage}[t]{1\linewidth}
\centering
\setlength{\tabcolsep}{1pt}
\footnotesize
\tablestyle{12pt}{1.2}
\resizebox{1.0\linewidth}{!}
{
\begin{tabular}{l|c|c}
method & $\operatorname{AP}_{25}$ & $\operatorname{AP}_{50}$ \\
\shline
Baseline (w/o RPE) & $71.4$ & $47.6$ \\
Baseline + CRPE (Stratrified Transformer)  &  $74.7$ & $58.1$ \\
Baseline + CRPE (EQNet) &  $73.1$ & $54.4$ \\
Baseline + Cond-CA & $74.7$ & $55.8$ \\
Baseline + DAB-CA  & $75.4$  & $56.0$ \\
 \rowcolor{gray!10}Baseline + 3DV-RPE  & $77.8$ & $66.0$ \\
\end{tabular}
}
\caption{\small
Comparison to other attention modulation methods.
We only change the decoder cross-attention scheme and keep all other settings the same for comparison fairness.
}
\label{tab:compare_crpe}
\end{minipage}
\end{table}

\begin{table}[t]
\begin{minipage}[t]{1\linewidth}
\vspace{-2mm}
\centering
\setlength{\tabcolsep}{5pt}
\footnotesize
\tablestyle{2pt}{1.2}
\resizebox{1.0\linewidth}{!}
{
\begin{tabular}{l|c|c|c|c|c}
method & $\#$ Scenes/second & Latency/scene & GPU Memory & $\operatorname{AP}_{25}$ & $\operatorname{AP}_{50}$ \\
\shline
FCAF3D & $7.8$ & $128$ms & $628$M & $71.5$  & $57.3$   \\
CAGroup3D & $2.1$ & $480$ms & $1138$M & $75.1$ & $61.3$   \\
Ours (light) &  $7.7$ & $130$ms & $489$M & $75.6$ & $62.7$   \\
Ours & $4.2$ & $240$ms & $642$M & $77.8$ & $66.0$   \\
\end{tabular}
}
\caption{\small
Inference cost comparison. We evaluate all numbers on a Tesla V100 PCIe 16 GB GPU with batch size as $1$ for a fair comparison.
}
\label{tab:compare_inference}
\vspace{1mm}
\end{minipage}
\end{table}

\vspace{1mm}
\noindent \textbf{Light-weight FFN.}
Table~\ref{tab:ablate_ffn} reports the comparison results on the effect of proposed light FFN.
According to the results, we observe that using the light-weight FFN significantly boosts the $\operatorname{AP}_{50}$ from $62.8$ to $65.0$, thus showing the advantages of using a set of adaptive predicted initial 3D bounding boxes over a set of pre-defined 3D bounding boxes of the same size.

\vspace{1mm}
\noindent \textbf{Number of points during training and evaluation.}
In Table~\ref{tab:ablate_point_num}, we report the comparison results when using different number of points during training. We observe that using $100$K points achieves consistently better performance, thus we choose $100$K points.

\vspace{1mm}
\noindent \textbf{One-to-many matching.}
Table~\ref{tab:effect_one2many} shows the comparison results when choosing different hyper-parameters for a one-to-many matching scheme.
For example, we find increasing the number of queries and the number of ground truth repeating times even hurts the performance when training with $40$K points but improves the performance when training with $100$K.

\vspace{1mm}
\noindent \textbf{Table shape.}
In Table~\ref{tab:ablate_shape}, we show the effect of different shapes for the pre-defined 3DV-RPE table. We find that $10\times10\times10$ achieves the best results. Our approach is less sensitive to the shape of the 3DV-RPE table thanks to the signed log function, which improves the interpolation quality to some degree.

\vspace{1mm}
\noindent \textbf{Comparison with other attention modulation methods.}
We summarize the comparison results with other advanced related methods including contextual relative position encoding (CRPE)~\cite{lai2022stratified,yang2022unified}, conditional cross-attention (Cond-CA)~\cite{meng2021CondDETR}, dynamic anchor box cross-attention (DAB-CA)~\cite{liu2022dab} in Table~\ref{tab:compare_crpe}.
We report the comparison results under the most strong settings, i.e., $540$ training epochs.
Accordingly, we see that (i) both CRPE (Stratified Transformer~\cite{lai2022stratified}) and CRPE (EQNet~\cite{yang2022unified}) consistently improve the baseline; (ii) our 3DV-RPE achieves the best performance.
The reason is that the CRPE methods of Stratified-Transformer~\cite{lai2022stratified} and EQNet~\cite{yang2022unified} only consider the center point of the 3D box while our 3DV-RPE explicitly considers the $8{\times}$ vertex points and rotated angle of the 3D box. Our method encodes the box size and the six faces, thus modeling the accurate position relations between all other points and the 3D bounding box (supported by the much larger gains on $\operatorname{AP}_{50}$).

\vspace{1mm}
\noindent \textbf{Inference complexity comparison.}
Table~\ref{tab:compare_inference} reports the comparison results to FCAF$3$D and CAGroup3D. Accordingly, our method achieves a better performance-efficiency trade-off than CAGroup3D.
We also provide a light version by decreasing the number of 3D object query from $1024$ to $256$.
Notably, the reported latency of CAGroup3D is close to the numbers in their \href{https://github.com/Haiyang-W/CAGroup3D#main-results}{official logs} but different from the numbers reported in the paper (179.3ms tested on RTX 3090 GPU). The authors of CAGroup3D have acknowledged this \href{https://github.com/Haiyang-W/CAGroup3D#todo}{issue} in their GitHub repository.

\section*{B. More Qualitative Results and Analysis}

\begin{figure*}[t]
\centering
\includegraphics[height=0.5\columnwidth]{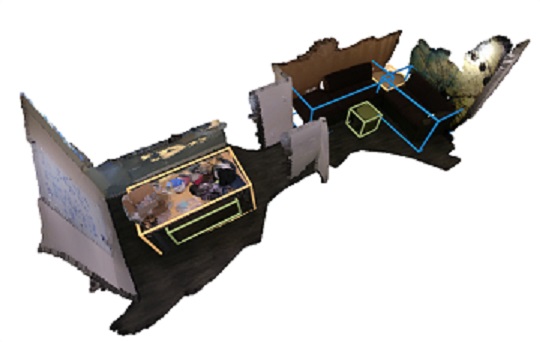}
\hspace{5pt}
\includegraphics[height=0.5\columnwidth]{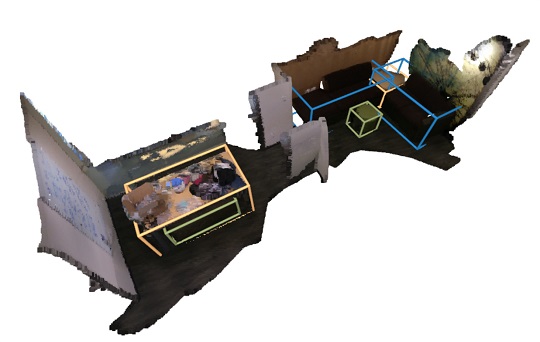} \\
\includegraphics[height=0.5\columnwidth]{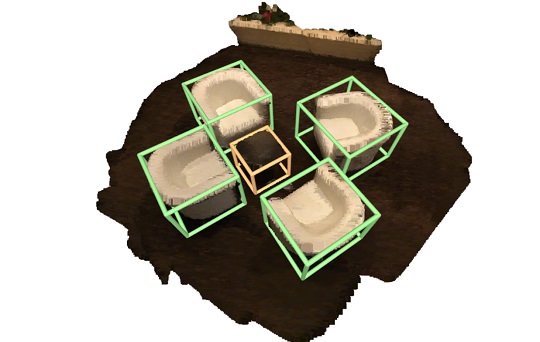}
\hspace{5pt}
\includegraphics[height=0.5\columnwidth]{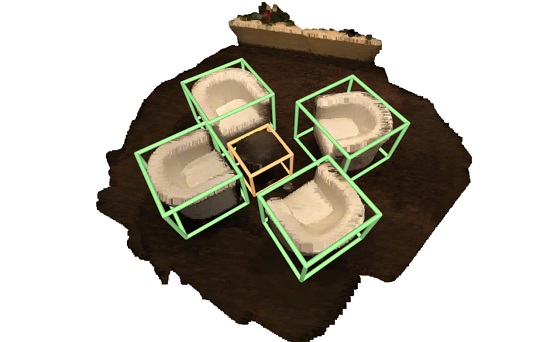} \\
\includegraphics[height=0.5\columnwidth]{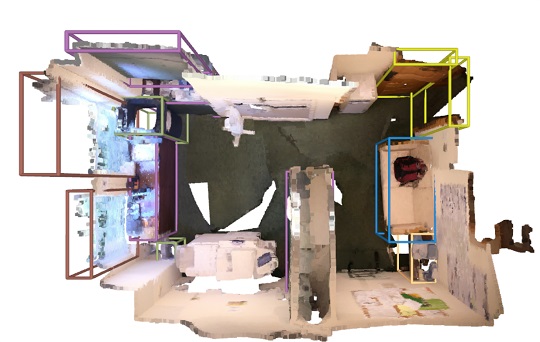}
\hspace{5pt}
\includegraphics[height=0.5\columnwidth]{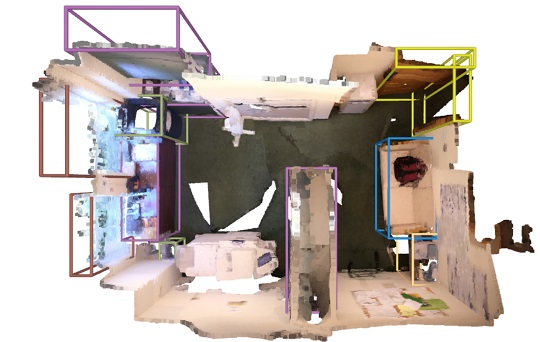} \\
\includegraphics[height=0.5\columnwidth]{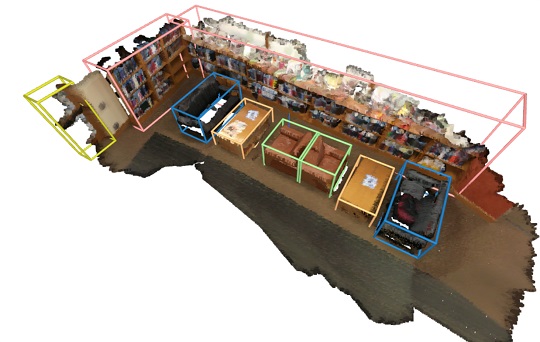}
\hspace{5pt}
\includegraphics[height=0.5\columnwidth]{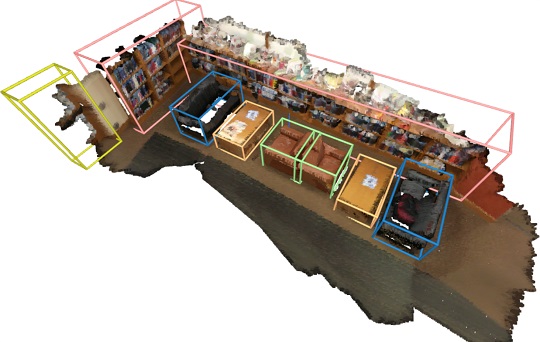} \\
\includegraphics[height=0.5\columnwidth]{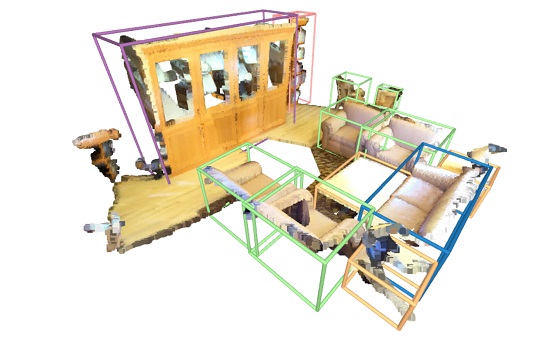}
\hspace{5pt}
\includegraphics[height=0.5\columnwidth]{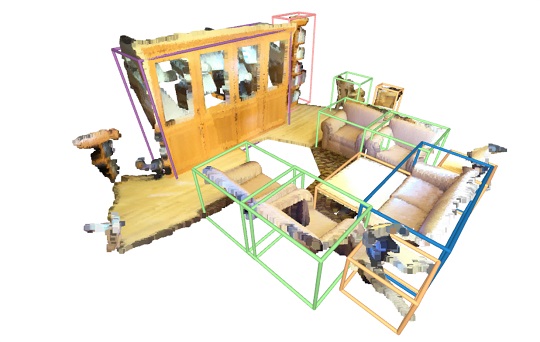} \\
\caption{\small{
More qualitative results of 3D object detection on ScanNetV2. The ground-truth is shown in the first column and our method's detection results are shown in the second column.
}}
\label{fig:vis_scannet}
\end{figure*}

\begin{figure*}[t]
\centering
\includegraphics[height=0.5\columnwidth]{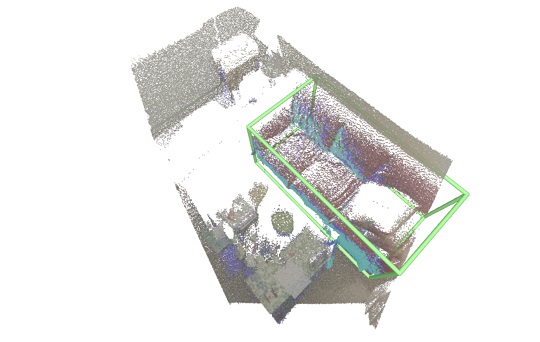}
\hspace{5pt}
\includegraphics[height=0.5\columnwidth]{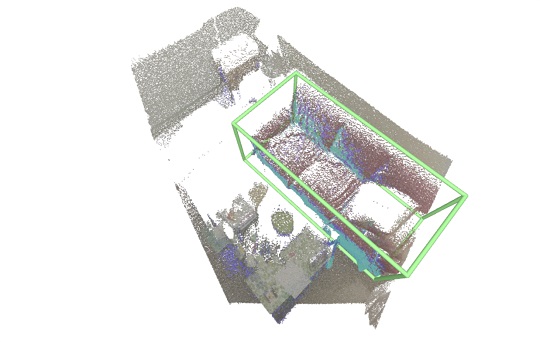} \\
\includegraphics[height=0.5\columnwidth]{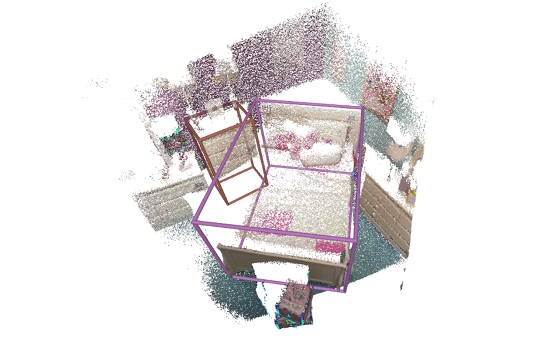}
\hspace{5pt}
\includegraphics[height=0.5\columnwidth]{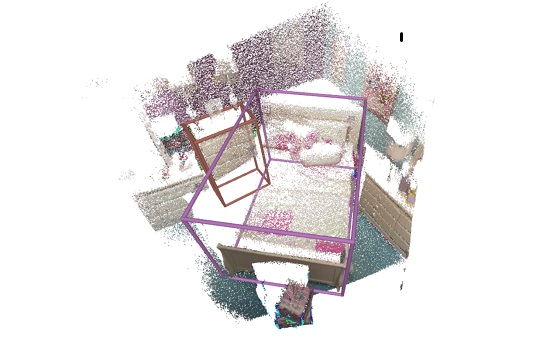} \\
\includegraphics[height=0.5\columnwidth]{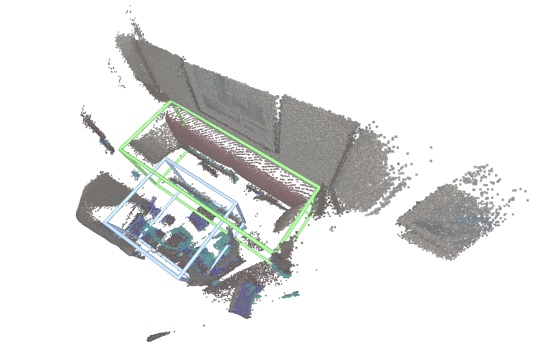}
\hspace{5pt}
\includegraphics[height=0.5\columnwidth]{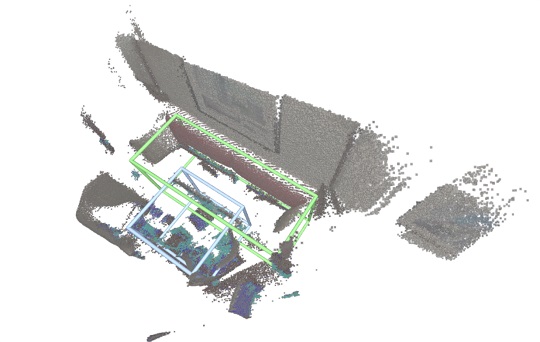} \\
\includegraphics[height=0.5\columnwidth]{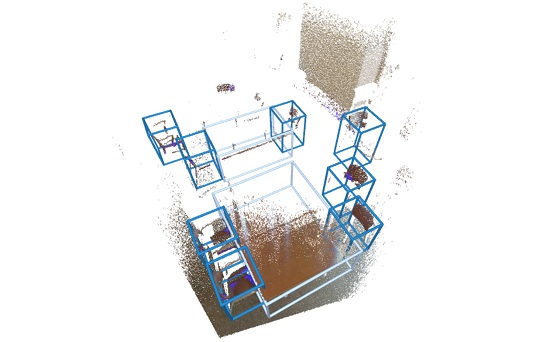}
\hspace{5pt}
\includegraphics[height=0.5\columnwidth]{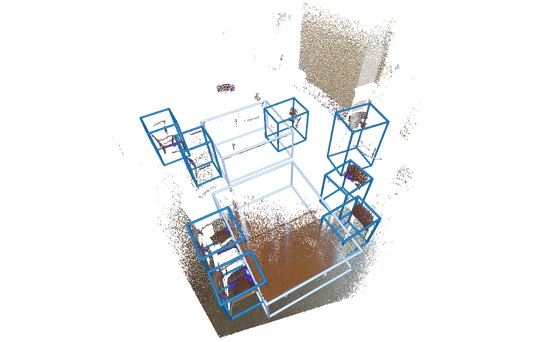} \\

\caption{\small{
More qualitative results of 3D object detection on SUN RGB-D. The ground truth is shown in the first column and our method's detection results are shown in the second column.
}}
\label{fig:vis_sunrgbd}
\end{figure*}

We show more qualitative examples of our V-DETR detection on ScanNet and SUN RGB-D in Figure~\ref{fig:vis_scannet} and Figure~\ref{fig:vis_sunrgbd}, respectively.
We can observe that our method can find most of the target objects in various scenes.

\begin{figure*}[t]
\centering
\includegraphics[height=0.25\columnwidth]{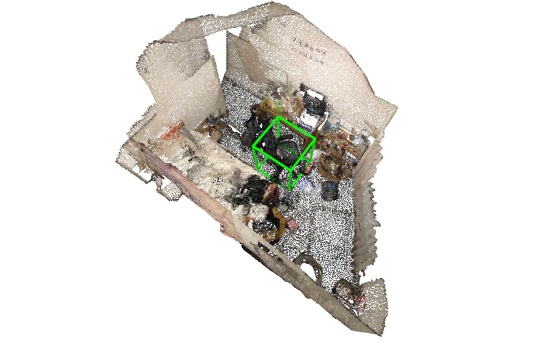}
\hfill
\includegraphics[height=0.25\columnwidth]{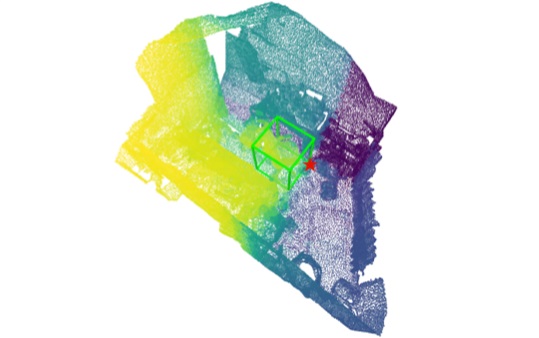}
\hfill
\includegraphics[height=0.25\columnwidth]{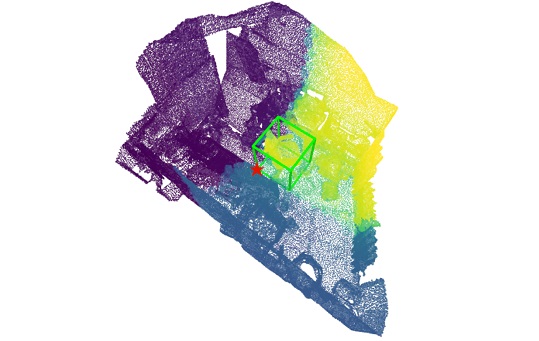}
\hfill
\includegraphics[height=0.25\columnwidth]{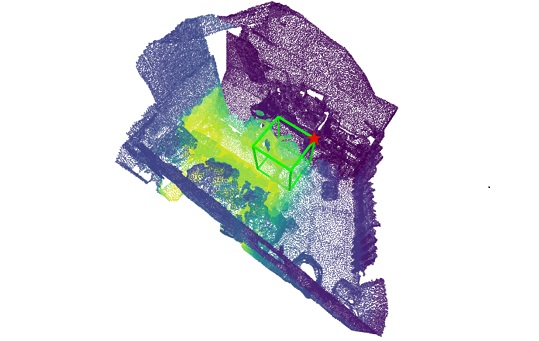}
\hfill
\includegraphics[height=0.25\columnwidth]{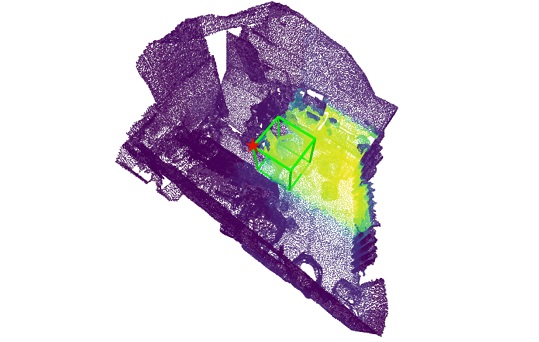}
\hfill
\\
\includegraphics[height=0.25\columnwidth]{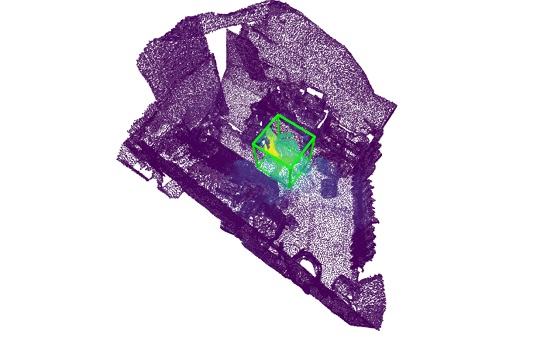} 
\hfill
\includegraphics[height=0.25\columnwidth]{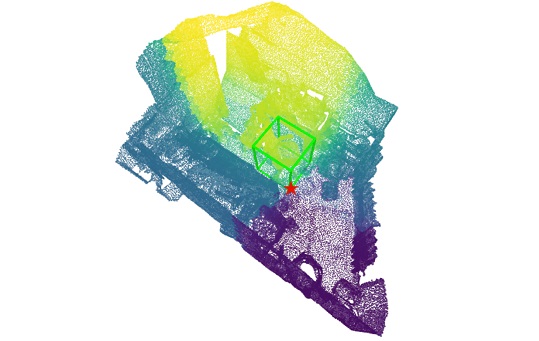}  
\hfill
\includegraphics[height=0.25\columnwidth]{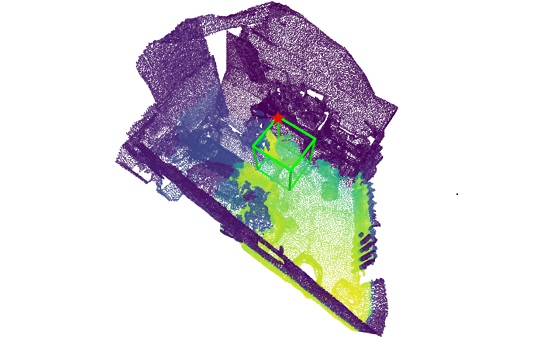}
\hfill
\includegraphics[height=0.25\columnwidth]{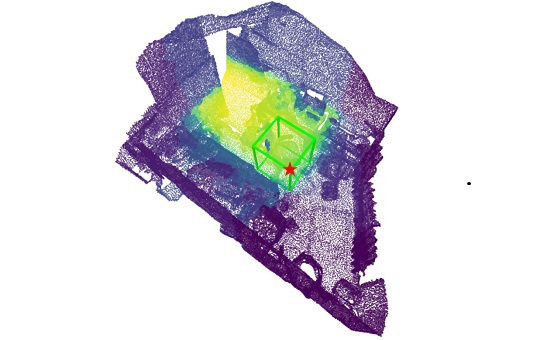}
\hfill
\includegraphics[height=0.25\columnwidth]{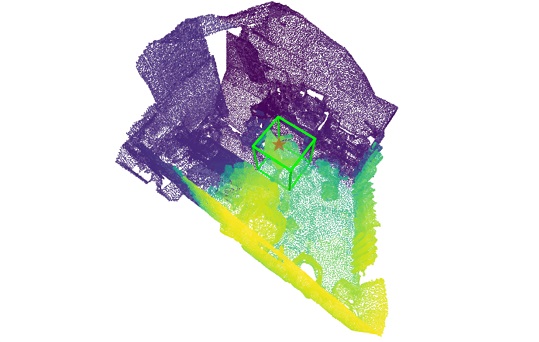}  \\
\vspace{10mm}

\includegraphics[height=0.26\columnwidth]{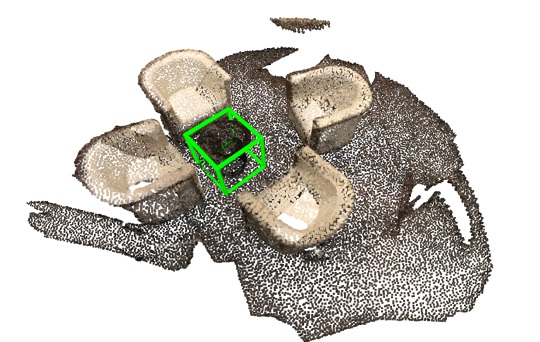}
\hfill
\includegraphics[height=0.26\columnwidth]{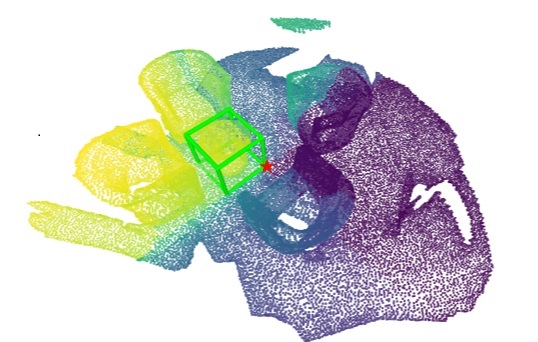}
\hfill
\includegraphics[height=0.26\columnwidth]{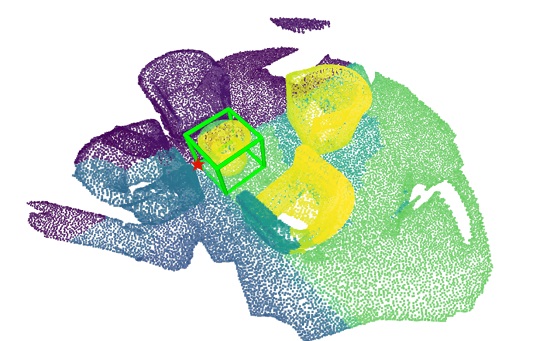}
\hfill
\includegraphics[height=0.26\columnwidth]{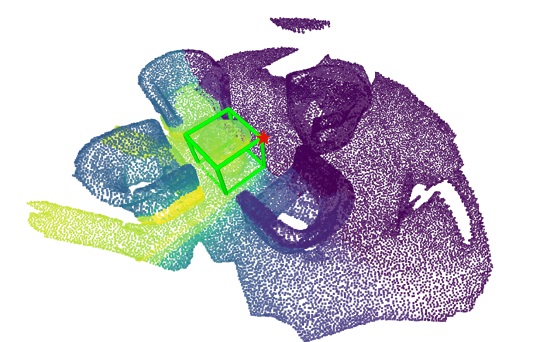}
\hfill
\includegraphics[height=0.26\columnwidth]{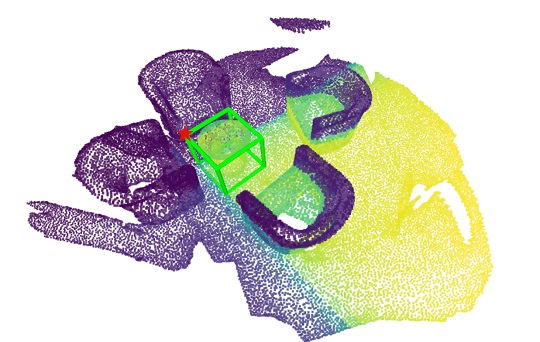}
\hfill
\\
\includegraphics[height=0.25\columnwidth]{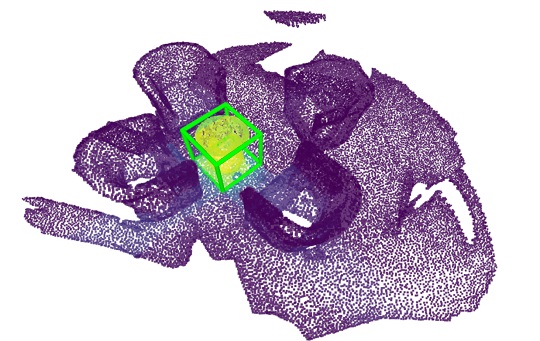} 
\hfill
\includegraphics[height=0.25\columnwidth]{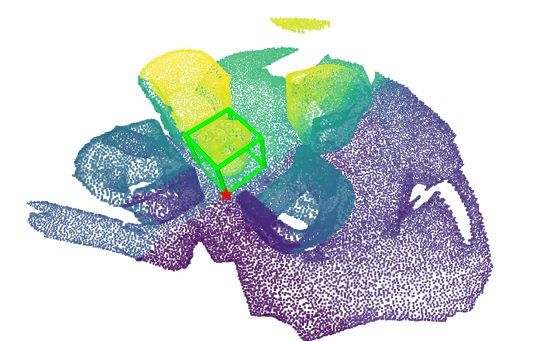}  
\hfill
\includegraphics[height=0.25\columnwidth]{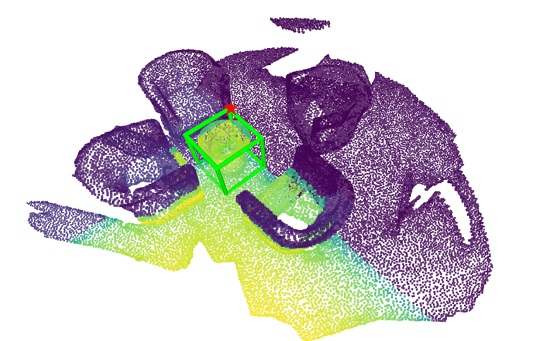}
\hfill
\includegraphics[height=0.25\columnwidth]{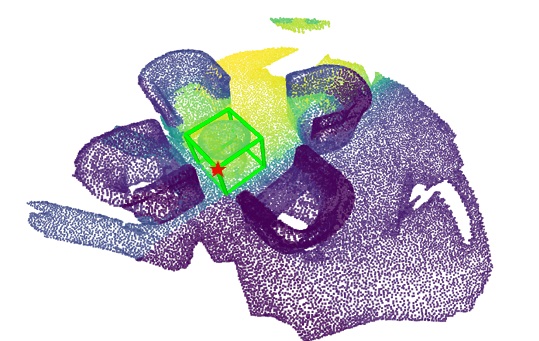}
\hfill
\includegraphics[height=0.25\columnwidth]{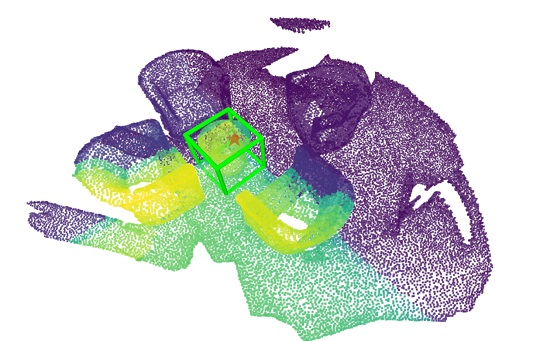}  \\
\vspace{10mm}
\includegraphics[height=0.25\columnwidth]{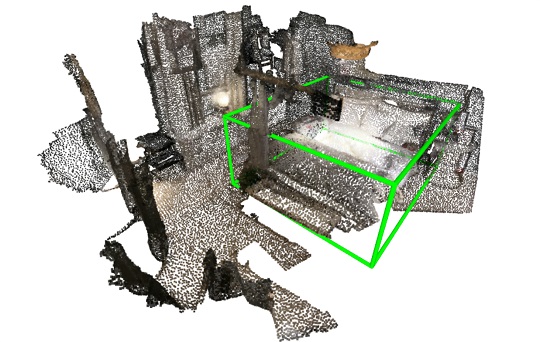}
\hfill
\includegraphics[height=0.25\columnwidth]{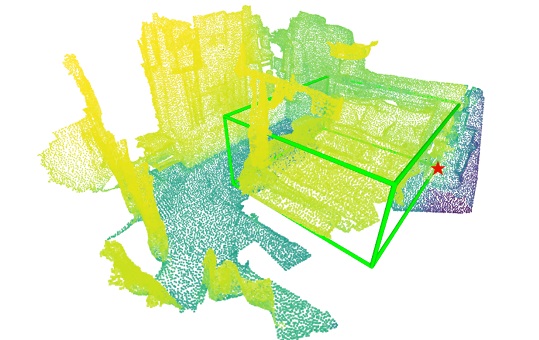}
\hfill
\includegraphics[height=0.25\columnwidth]{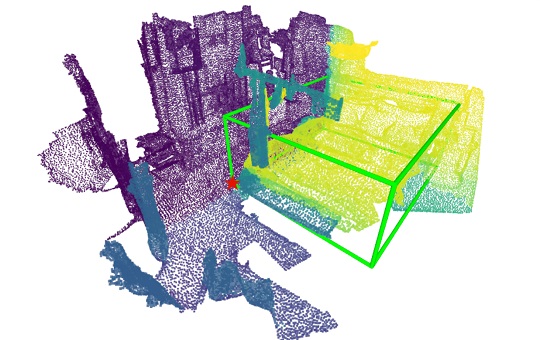}
\hfill
\includegraphics[height=0.25\columnwidth]{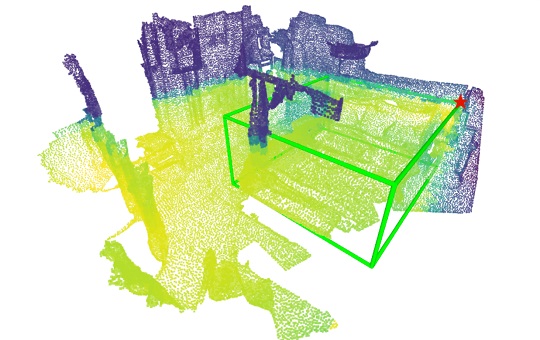}
\hfill
\includegraphics[height=0.25\columnwidth]{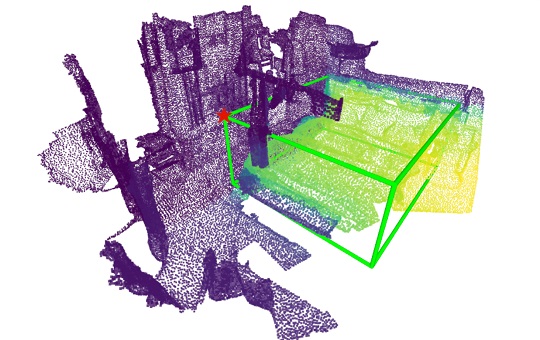}
\hfill
\\
\includegraphics[height=0.25\columnwidth]{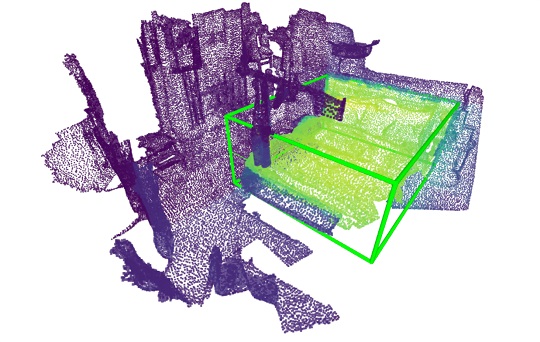} 
\hfill
\includegraphics[height=0.25\columnwidth]{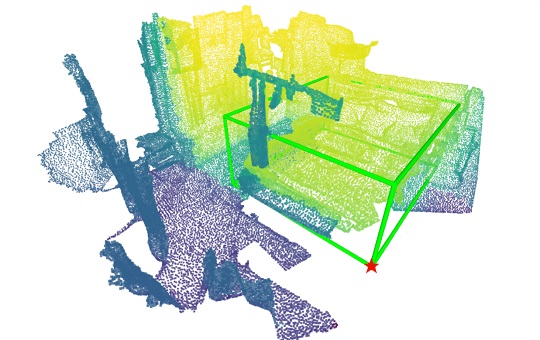}  
\hfill
\includegraphics[height=0.25\columnwidth]{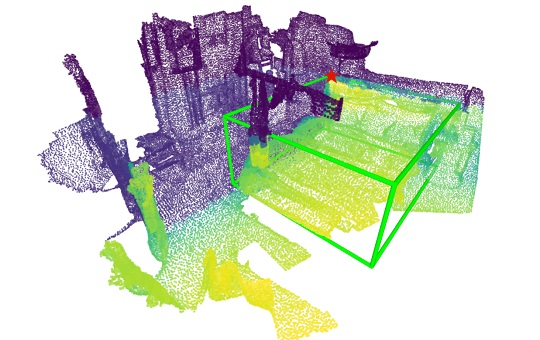}
\hfill
\includegraphics[height=0.25\columnwidth]{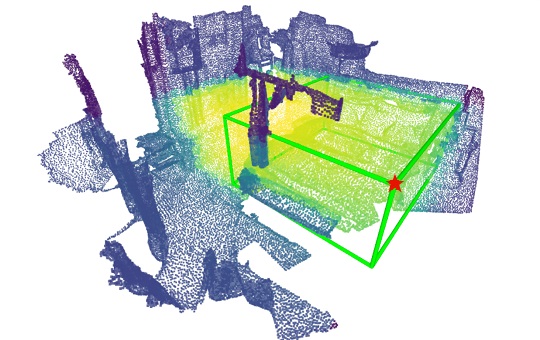}
\hfill
\includegraphics[height=0.25\columnwidth]{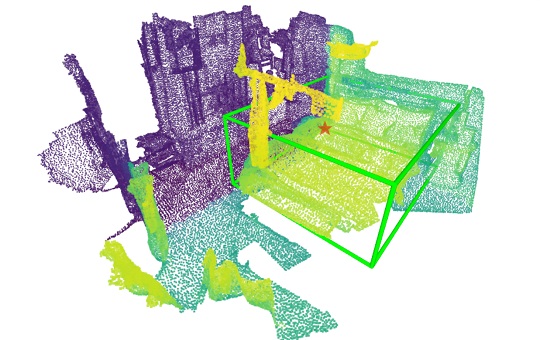}  \\
\vspace{5mm}
\caption{\small{
\textbf{Illustration of the spatial attention maps learned by our 3DV-RPE on ScanNetV2 scenes.} Each scene consists of two rows. We draw a green cube to mark the detected 3D bounding box and a red star at its eight vertices. We average the head dimension of each $\mathbf{P}_i$ and show the spatial cross-attention maps for eight vertices (columns $2$-$5$). Column $1$ shows the input scene and the merged attention maps. The color shows the attention values: yellow is high and blue is low. We see that (i) each vertex's attention map highlights the regions inside the cube from that vertex, and (ii) the combined attention maps focus on the regions inside the red cubes.}}
\label{fig:3dvrpe_visual_exp}
\end{figure*}

Figure~\ref{fig:3dvrpe_visual_exp} shows the spatial cross-attention maps of our 3DV-RPE on three ScanNetV2 scenes. We see that (i) our 3DV-RPE can find the 3D bounding boxes accurately and (ii) each vertex's RPE can enhance the regions inside the boxes from that vertex.


\end{document}